%% file: tmlr.tex
\documentclass[10pt]{article}
\usepackage[preprint]{tmlr}

\usepackage{ragged2e}
\usepackage{xcolor}
\usepackage{tcolorbox}
\tcbuselibrary{skins,breakable}
\usepackage{url}
\usepackage{subcaption}
\usepackage{listings}
\usepackage{booktabs}
\usepackage{multirow}
\usepackage{makecell}
\usepackage{array}
\usepackage{caption}
\usepackage{wrapfig}
\usepackage{graphicx}

\usepackage{algorithm}
\usepackage{algpseudocode}
\usepackage{amsmath}

\usepackage{enumitem}
\newenvironment{itemize*}%
 {\leftmargini=20pt\begin{itemize}%
  \setlength{\itemsep}{3pt}%
  \setlength{\parskip}{0pt}%
  }%
 {\end{itemize}} 
\newenvironment{enumerate*}%
 {\begin{enumerate}%
  \setlength{\itemsep}{0pt}%
  \setlength{\parskip}{0pt}}%
 {\end{enumerate}}

\DeclareTextCommand{\textquotedbl}{OT1}{\char`\"}
\lstdefinestyle{jsonTiny}{
  basicstyle=\ttfamily\scriptsize,
  breaklines=true,
  breakindent=0pt,
  columns=fullflexible,
  keepspaces=true,
  showstringspaces=false,
  upquote=true,
  frame=none
}

\usepackage[
  colorlinks=true,
  linkcolor=tmlrBlueDark,
  citecolor=tmlrCiteBlue,
  urlcolor=tmlrBlueDark
]{hyperref}
\usepackage[capitalize,noabbrev]{cleveref}

\graphicspath{{figures/}}
\newcommand{\setting}[1]{\par\noindent\textbf{Setting.} #1\par}
\newcommand{\result}[1]{\par\noindent\textbf{Results.} #1\par}
\newcommand{\insight}[1]{\par\noindent\textbf{Insight.} #1\par}
\newcommand{\better}[1]{\textcolor{tmlrBlueDark}{#1}}
\newcommand{\worse}[1]{\textcolor{tmlrBlue}{#1}}

\input{vals/macros}
\title{AI4AI at Test-Time: Strong-to-Weak Capability Transfer via Harnesses}
\long\def\paperabstract{Recent work on distillation transfers the capabilities of large models to smaller ones often by updating the latter's parameters, through teacher forcing, on-policy distillation, and related training-time methods. In this paper, we ask whether such transfer can instead occur at test time. We study \textbf{strong-to-weak scaffolding}: whether a stronger \emph{builder model} can construct inference-time harnesses that help a weaker \emph{target model} solve tasks more reliably without any parameter updates. Using four representative Theory-of-Mind benchmarks, each builder model uses 5\% of the data as a validation set to iteratively refine its harness over multiple rounds, after which the finalized harness is evaluated on the full test set. Empirically, this form of test-time capability transfer is highly effective, nearly doubling average target-model performance from 0.49 to 0.91. Our analysis shows that the gains come primarily from offloading unstable model reasoning into deterministic code, benchmark-specific routing, and strict answer-format enforcement, rather than from encouraging the target model to reason more extensively or sample more broadly. We further find that builder-model reasoning effort improves harness quality monotonically, platform effects are modest relative to the builder model's own capability, and weaker target models receive the largest gains. These results suggest that inference-time harness design is an important complement to conventional training-time distillation, enabling strong models to transfer cognitive structure to weaker models without retraining.}

\author{Cheng Qian$^{1,2}$, Wenting Zhao$^{1}$, Liangwei Yang$^{1}$, Heng Wang$^{1,2}$, Jielin Qiu$^{1}$, Heng Ji$^{2}$, Silvio Savarese$^{1}$,\\
\textbf{Huan Wang$^{1}$, Shelby Heinecke$^{1}$}\vspace{1.5mm}\\
$^1$Salesforce AI Research, $^2$University of Illinois Urbana-Champaign \vspace{1mm}\\
}

\def\month{MM}
\def\year{YYYY}
\def\openreview{\url{https://openreview.net/forum?id=XXXX}}

\begin{document}

\maketitle

\input{sections/intro}

\input{sections/related_works}
\input{sections/method}
\input{sections/setup}
\input{sections/overview}

\input{sections/a0_main}
\input{sections/a1_stability}
\input{sections/a2_refine}
\input{sections/a3_techniques}
\input{sections/a4_platform}
\input{sections/a5_weak}
\input{sections/a6_effort}
\input{sections/a7_attribution}
\input{sections/a8_cogload}
\input{sections/a9_errors}
\input{sections/conclusion}

\bibliography{tmlr}
\bibliographystyle{tmlr}

\input{sections/appendix}

\end{document}

%% file: vals/macros.tex
\newcommand{\vaFiveGemBigtomShare}{96}
\newcommand{\vaFiveGemBigtomUplift}{0.42}
\newcommand{\vaFiveGemHitomUplift}{-0.04}
\newcommand{\vaFiveGemMumaUplift}{-0.02}
\newcommand{\vaFiveGemRegCells}{9}
\newcommand{\vaFiveGptRegCells}{0}
\newcommand{\vaFiveHeadroomCorr}{0.75}
\newcommand{\vaFiveNBuilders}{5}
\newcommand{\vaFiveRegCellsTotal}{20}
\newcommand{\vbaseGeminiMacro}{0.761}
\newcommand{\vbaseGptMacro}{0.488}
\newcommand{\vbestRunAvg}{0.912}
\newcommand{\vbestRunBuilder}{GPT-5.5}
\newcommand{\vbestRunPlatform}{GPT Codex}
\newcommand{\vbestRunRelUplift}{87}
\newcommand{\vbestRunUplift}{0.423}
\newcommand{\vcorrNvalFull}{0.17}
\newcommand{\vcorrValBestFull}{0.96}
\newcommand{\veffortHighMean}{0.807}
\newcommand{\veffortLowMean}{0.711}
\newcommand{\veffortMedMean}{0.793}
\newcommand{\veffortSpearman}{0.77}
\newcommand{\veffortXhighHighP}{0.013}
\newcommand{\veffortXhighLowP}{0.002}
\newcommand{\veffortXhighMean}{0.856}
\newcommand{\vfracBeatBase}{100\%}
\newcommand{\vhitomOrderFour}{0.700}
\newcommand{\vhitomOrderZero}{0.999}
\newcommand{\vmaxNVal}{15}
\newcommand{\vmaxWithinRange}{0.201}
\newcommand{\vmcnemarGptBroke}{105}
\newcommand{\vmcnemarGptFixed}{1717}
\newcommand{\vmeanFracBroke}{7\%}
\newcommand{\vmeanFracFixed}{83\%}
\newcommand{\vmeanNVal}{4.9}
\newcommand{\vmeanScaffoldGpt}{0.763}
\newcommand{\vmeanUpliftGpt}{0.275}
\newcommand{\vmeanValFullGap}{0.021}
\newcommand{\vmeanValGain}{0.216}
\newcommand{\vmeanWithinSd}{0.036}
\newcommand{\vmedianNVal}{5}
\newcommand{\vminNVal}{2}
\newcommand{\vnCells}{11}
\newcommand{\vnRunsGpt}{57}
\newcommand{\vnRunsTotal}{72}
\newcommand{\vnTopScaffolds}{8}
\newcommand{\voracleFixCov}{97\%}
\newcommand{\vplatEffortHighDelta}{+0.038}
\newcommand{\vplatEffortLowDelta}{-0.034}
\newcommand{\vplatEffortMedDelta}{+0.045}
\newcommand{\vplatEffortXhighDelta}{+0.032}
\newcommand{\vplatMargClaudecode}{0.799}
\newcommand{\vplatMargCursor}{0.750}
\newcommand{\vplatMargGptcodex}{0.754}
\newcommand{\vplatNativeAllDelta}{+0.013}
\newcommand{\vplatNativeClaudeDelta}{+0.008}
\newcommand{\vplatNativeGptDelta}{+0.020}
\newcommand{\vplatNativeNBetter}{5}
\newcommand{\vplatNativeNTotal}{8}
\newcommand{\vplatNativePairedP}{0.484}
\newcommand{\vuserHarnessGeminiMacro}{0.941}
\newcommand{\vuserHarnessGptMacro}{0.939}
\newcommand{\vweakGeminiBase}{0.761}
\newcommand{\vweakGeminiScaf}{0.871}
\newcommand{\vweakGeminiUplift}{0.110}
\newcommand{\vweakGptBase}{0.488}
\newcommand{\vweakGptScaf}{0.750}
\newcommand{\vweakGptUplift}{0.262}

%% file: sections/intro.tex
\section{Introduction}
\label{sec:intro}

Recent progress in model distillation has made it increasingly plausible to deploy smaller language model experts in settings that once required much larger ones~\citep{hinton2015distilling,hsieh2023distilling,agarwal2024on}. Most existing approaches transfer capability by changing the weak model itself. For instance, \textit{data distillation} trains a small student on examples, rationales, or demonstrations produced by a stronger teacher. \textit{On-policy distillation} further exposes the student to dense feedback, preferences, or reward signals while it acts, allowing the student to internalize behaviors that would otherwise be difficult to acquire from static data alone~\citep{agarwal2024on,ouyang2022training}. These approaches are effective, but they share a common premise: closing the gap between a strong model and a weak model needs additional training.

This paper studies a complementary premise. When a small model fails on a task, the failure may reflect not only insufficient internal capability, but also excessive \textit{cognitive load} imposed by the way the task is presented.~\citep{sweller1988cognitive}. A system can therefore improve performance in two ways: it can make the model more capable, or it can make the task easier for the model to solve. The first route is the dominant route of distillation. The second route is increasingly realized through \textbf{inference-time harnesses}: external structures that surround a model with routing logic, prompt templates, verification checks, memory, and tool use~\citep{wei2022chain,yao2023react,schick2023toolformer,madaan2023self}. Rather updating the target model's parameters, a harness engineers the conditions under which the target model reasons.

In this paper, we investigate into this \textbf{strong-to-weak scaffolding}. Specifically, a strong \emph{builder} model is asked to construct a scaffold: any combination of task routing, prompt templates, deterministic solvers, few-shot exemplars, verification passes, or format enforcement designed to improve a fixed weaker \emph{target} model on a hidden test set. The target model is not trained, and the builder never observes the full test set; its only opportunity to improve performance is to design an inference-time environment that transfers across examples. Thus, a successful scaffold must capture reusable skills and task structure rather than memorize instance-specific answers. This setting isolates a practical form of strong-to-weak transfer in which capability is transferred not through model weights, but through the harness that shapes how the weak model receives, reasons, and responds.

Strong-to-weak scaffolding is becoming more important as model deployment shifts from single prompts to agentic systems~\citep{wang2024survey,ke2025survey}. In practice, small models are rarely used in isolation. They are embedded in pipelines that parse inputs, select tools, check answers, and decompose tasks~\citep{yue2026from}. Yet we still lack a systematic account of why these harnesses help, when they are stable, which design choices matter, and how much of the improvement reflects genuine reasoning support rather than benchmark-specific shortcuts~\citep{ullman2023large,riemer2025position}. Without such an account, harness engineering remains difficult to compare, reproduce, and improve.

We investigate these questions in the domain of Theory-of-Mind (ToM) reasoning. ToM benchmarks are a useful stress test because they require models to track nested beliefs, perspective shifts, hidden information, and Bayesian goal inference~\citep{chen2025theory,he2023hitom,baker2009action}. These demands are challenging for smaller models, but they also contain structure that an external scaffold may exploit: tasks can often be routed by subtype, decomposed into intermediate states, checked for consistency, or solved partly through symbolic procedures~\citep{zhang2025autotom}. To fully understand the effect of scaffolding, we analyze a large corpus of strong-builder, weak-target runs and investigate into the following aspects:
\begin{itemize}[topsep=-3pt, partopsep=-3pt, leftmargin=*, itemsep=-3pt]
    \item \textbf{Effect size:} how much scaffolding improves weak target model's accuracy overall;
    \item \textbf{Stability:} whether independently built scaffolds produce consistent gains;
    \item \textbf{Validation effort:} how much times the validation data is used, and whether using more helps;
    \item \textbf{Scaffold techniques:} what specific mechanisms builders actually implement;
    \item \textbf{Platform effects:} if the builder's own agentic harness changes outcomes;
    \item \textbf{Target dependence:} how scaffolding strategies and gains vary with the weak target model;
    \item \textbf{Builder reasoning effort:} whether stronger internal reasoning of builder improves scaffolds;
    \item \textbf{Causal mechanisms:} which scaffold features are associated with accuracy gains;
    \item \textbf{Cognitive-load reduction:} how much reasoning is shifted away from the target model;
    \item \textbf{Failure modes:} where even the best scaffolds continue to make errors.
\end{itemize}

Our empirical results show that strong-to-weak scaffolding is both large and robust. The best scaffold raises GPT-5.4-mini from a macro-average accuracy of $0.49$ to $0.91$, an absolute gain of $0.42$, and every builder configuration yields positive net uplift. The gains are driven less by using more validation data, sampling more, or simply eliciting longer reasoning, and more by structure externalization, deterministic offloading, and strict format enforcement, etc.

Besides, the improvements are not uniform in kind. On BigToM, the best scaffolds discover that answers can be fully exploited through structured reasoning, thus turning the benchmark into compilable skills and rules. On the other benchmarks, the gains reflect genuine reduction of reasoning burden rather than a shortcut. Residual errors concentrate in the hardest regimes, especially higher-order Hi-ToM cases with recursion depth at least two and Bayesian goal-inference subtypes. We also find that builder reasoning effort improves scaffold quality monotonically across effort tiers, platform effects are modest relative to builder-model effects, and scaffolding helps the weaker GPT-5.4-mini target more than the already stronger Gemini-3.5-flash target. In summary, this study makes three contributions:
\begin{itemize}[topsep=-3pt, partopsep=-3pt, leftmargin=*, itemsep=-3pt]
\item First, we formalize \emph{strong-to-weak scaffolding} as a distinct inference-time capability transfer setting.
\item Second, we provide a systematic empirical analysis of its effect size, stability, validation efficiency, platform and target dependence, mechanisms, and cognitive-load reduction.
\item Third, we identify actionable design principles: successful scaffolds often rely on deterministic offloading, benchmark-aware routing, format control, and targeted decomposition than brute-force validation search.
\end{itemize}

Looking ahead, strong-to-weak scaffolding is valuable not only as a deployment strategy but also as a way to evaluate builder-model capability. For deployment, it offers a practical route to improving weaker or cheaper models at inference time, without changing their weights. This closely aligns with the motivation behind automatic agent-harness self-evolution. For evaluation, it reframes the question from ``How well can a model solve a task?'' to ``How well can a stronger model construct the conditions under which a weaker model can solve it?'' Given the same hidden task, validation budget, and open-ended workspace, the quality of the resulting scaffold becomes a measure of the builder's ability to externalize reasoning into procedures, tools, feedback loops, and constraints. More broadly, this setting creates a natural platform for studying harness evolution itself: what structures builders invent, where their designs fail, how they revise them, and which forms of external organization most efficiently translate strong-model insight into weak-model performance.

%% file: sections/related_works.tex
\section{Related Work}

\textbf{Capability transfer and distillation.}
A line of existing work studies how capabilities of a larger or stronger model can be transferred to a smaller or cheaper one through training. Classical knowledge distillation compresses an ensemble or high-capacity teacher into a deployable student by training the student to match softened teacher outputs \citep{hinton2015distilling}. Recent language-model distillation methods extend this paradigm by transferring rationales, traces, or task-specific reasoning supervision: for example, distilling step-by-step uses teacher-generated rationales as additional supervision for smaller task models \citep{hsieh2023distilling}, while on-policy distillation trains on student-generated sequences with teacher feedback to reduce the distribution mismatch between training and inference \citep{agarwal2024on}. Instruction tuning and RLHF similarly alter the model policy through supervised and preference-based training signals \citep{ouyang2022training}. Related alignment work on weak-to-strong generalization asks whether weak supervision can elicit capabilities from a stronger model, but still studies capability transfer through model updating rather than through the inference environment \citep{burns2023weak}. Our work differs from these previous paradigms as a complementary test-time capability-transfer paradigm: instead of changing the weak target model's parameters, it asks whether a strong builder can externalize transferable task structure into a reusable harness that improves a weak target at inference time.

\textbf{Inference-time reasoning, prompting, and decomposition.}
A second line of work improves model reasoning without conventional fine-tuning by changing the inference procedure. Chain-of-thought prompting elicits intermediate reasoning steps from large models \citep{wei2022chain}, self-consistency improves reliability by sampling multiple reasoning paths and marginalizing over their answers \citep{wang2023selfconsistency}, and least-to-most prompting decomposes hard problems into easier subproblems solved sequentially \citep{zhou2023least}. Decomposed prompting generalizes this modular view by delegating subproblems to specialized prompts, models, or symbolic functions \citep{khot2023decomposed}. Iterative refinement methods such as Self-Refine use model-generated feedback to improve an initial answer over multiple rounds \citep{madaan2023self}, while Tree-of-Thoughts and Graph-of-Thoughts treat reasoning as search over structured intermediate states rather than as a single left-to-right trace \citep{yao2023tree,besta2024graph}. These methods show that test-time structure can substantially improve reasoning, but they typically optimize how the same model reasons on each instance. Our work is beyond single-model prompting by studying a cross-model scaffold-building setting in which a strong builder constructs a persistent inference-time procedure that a separate weaker target can execute across hidden examples.

\textbf{Tool use, programmatic reasoning, and deterministic offloading.}
Another closely related perspective treats reasoning failures as failures of execution, verification, or state tracking rather than failures of language understanding alone. Toolformer trains language models to decide when and how to call external APIs \citep{schick2023toolformer}, and ReAct interleaves natural-language reasoning with environment actions or tool calls \citep{yao2023react}. Program-aided methods such as PAL and Program-of-Thoughts ask the model to translate a problem into executable code, leaving exact computation to a Python interpreter \citep{gao2023pal,chen2023program}. Faithful chain-of-thought similarly separates translation from solving by using a deterministic solver to execute symbolic reasoning chains \citep{lyu2023faithful}. This literature motivates the view that a model can be made more reliable by moving fragile cognitive work into external tools, code, or checkable representations. Our work follows these motivations as a systematic empirical study of when such offloading emerges automatically from strong-builder scaffold design, showing that deterministic solvers, benchmark routing, and answer-format enforcement can transfer cognitive structure to a weaker target more effectively than merely encouraging longer target-model reasoning.

\textbf{Harness engineering and automated scaffold construction.}
Recent work increasingly treats the system surrounding an LLM, such as prompts, tools, memory, context management, execution interfaces, routing, and validation, as a first-class object of optimization. DSPy formalizes LM pipelines as declarative modules and compiles prompts or demonstrations against task metrics \citep{khattab2024dspy}. SWE-agent shows that the agent-computer interface itself can substantially affect coding-agent performance \citep{yang2024sweagent}. More recent work makes scaffold design itself an optimization target: ADAS frames agentic systems as code-defined artifacts that can be discovered by a meta-agent \citep{hu2025automated}, Meta-Harness searches over harness code using prior candidates, traces, and scores \citep{lee2026metaharness}, and Harness-Bench evaluates how harness configurations affect realistic agent workflows under shared environments and budgets \citep{yao2026harnessbench}. Survey work on code as agent harness further argues that code is becoming the operational substrate for state, verification, tool use, and feedback-driven control in agentic systems \citep{ning2026code}. Our work positions itself within this harness-engineering literature but isolates a specific strong-to-weak transfer regime, measuring how builder capability, builder reasoning effort, platform choice, validation budget, target-model headroom, and deterministic offloadability jointly shape the harness quality.

\textbf{Theory-of-Mind evaluation and mental-state scaffolds.}
Theory-of-Mind benchmarks provide a demanding testbed for scaffolding because they require tracking observations, beliefs, intentions, hidden information, and nested perspectives. BigToM procedurally generates social-reasoning evaluations from causal templates and shows that strong models may partially mirror human inference patterns while remaining unreliable \citep{gandhi2023bigtom}. Hi-ToM emphasizes higher-order recursive belief reasoning and finds that LLM performance declines as recursion depth increases \citep{he2023hitom}. MMToM-QA evaluates belief and goal inference in household activity settings and introduces Bayesian inverse planning accelerated by language models \citep{jin2024mmtom}, while MuMA-ToM extends ToM evaluation to embodied multi-agent interactions with goals, beliefs, and beliefs about others' goals \citep{shi2024muma}. Broader ToM evaluations such as ToMBench expand coverage across social-cognitive abilities \citep{chen2024tombench}, and UserHarness shows that explicit reconstruction of user beliefs, intentions, observations, and actions can provide a strong human-designed ToM harness \citep{qian2026userharness}. Our work does not intend as a new ToM benchmark or a manually designed ToM solver, but as a meta-evaluation of whether strong models can automatically discover reusable ToM scaffolds for weaker models and of where such scaffolds still fail when belief recursion or Bayesian goal inference resists compilation into rules and skills.

%% file: sections/method.tex
\section{Method}
\label{sec:method}

\begin{figure*}[!t]
    \centering
    \vspace{-0mm}
    \includegraphics[width=0.82\linewidth]{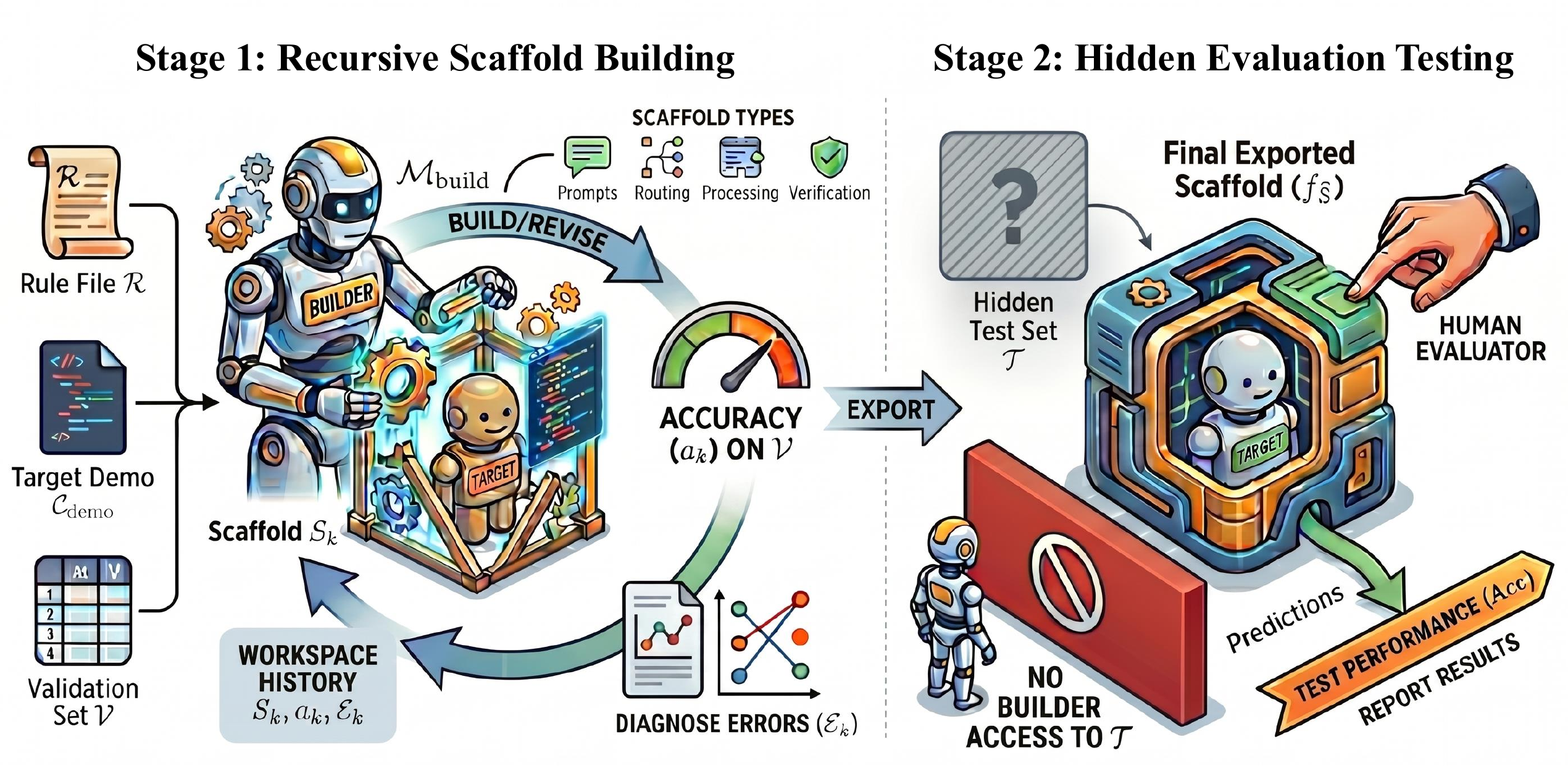}
    \caption{Overview of the \textbf{Strong-to-Weak Scaffolding} evaluation framework. During \textit{recursive scaffold building}, the builder model iteratively refines a scaffold to improve target-model performance on validation sets. During \textit{hidden evaluation testing}, the final scaffold is evaluated on the full hidden test set to measure target-model performance.}
    \label{fig:method}
    \vspace{0mm}
\end{figure*}

We study an automatic harness-building setting in which a strong \emph{builder} model constructs an inference-time scaffold for a fixed weaker \emph{target} model. Let $M_{\text{tar}}$ denote the target model and let $M_{\text{build}}$ denote the builder model. For each benchmark $\mathcal{D}^{(j)}$, we randomly sample a small validation split $\mathcal{V}^{(j)}\subset \mathcal{D}^{(j)}$ containing $5\%$ of the benchmark examples, and reserve the remaining examples as a hidden test split $\mathcal{T}^{(j)}$. We write
\[
\mathcal{V}=\bigcup_j \mathcal{V}^{(j)}, 
\qquad
\mathcal{T}=\bigcup_j \mathcal{T}^{(j)} .
\]
The builder has access only to $\mathcal{V}$ during scaffold construction, while $\mathcal{T}$ is held out and evaluated separately after the build process.
At the beginning of each run, the builder is placed inside an existing agentic coding harness $\mathcal{H}_{\text{build}}$ and given an initial workspace
\[
\mathcal{W}_0=\{\mathcal{R},\mathcal{C}_{\text{demo}},\mathcal{V}\},
\]

\input{tables/algorithm}

where $\mathcal{R}$ is a rule file describing the task instructions and submission format, $\mathcal{C}_{\text{demo}}$ is a demonstration file showing how to call $M_{\text{tar}}$, and $\mathcal{V}$ is the labeled validation set. The builder is not constrained to a fixed scaffold architecture. It may implement any inference-time procedure, including prompt templates, benchmark routing, deterministic pre- or post-processing, answer-format enforcement, verification passes, few-shot retrieval, or direct symbolic solvers. The only requirement is that the final scaffold exposes an entry point that can be applied to unseen test examples.

Conceptually, the builder searches over a space of possible scaffolds $\mathcal{S}$:
\[
S^\star
=
\arg\max_{S\in\mathcal{S}}
\operatorname{Acc}\bigl(S,M_{\text{tar}};\mathcal{T}\bigr),
\]
but since $\mathcal{T}$ is hidden, the builder can only use validation performance as a proxy:
\[
\hat{S}
=
\arg\max_{S\in\mathcal{S}_{\text{build}}}
\operatorname{Acc}\bigl(S,M_{\text{tar}};\mathcal{V}\bigr).
\]
A successful scaffold must therefore identify reusable task structure from the validation slice and transfer it to the hidden test set. After the builder finishes, a human evaluator runs the exported entry point on $\mathcal{T}$ without further builder intervention. Please refer to \Cref{algo:strong_to_weak_scaffold} for more details.

%% file: tables/algorithm.tex
\begin{wraptable}{r}{0.5\textwidth}
\vspace{-0mm}
\small
\captionof{algorithm}{Strong-to-weak scaffold building algorithm.}
\vspace{-1mm}
\label{algo:strong_to_weak_scaffold}

\noindent\rule{\linewidth}{0.6pt}
\vspace{-3mm}
\begin{algorithmic}[1]
\Require Builder model $M_{\mathrm{build}}$, target model $M_{\mathrm{tar}}$
\Require Builder-side harness $\mathcal{H}_{\mathrm{build}}$
\Require Rule file $\mathcal{R}$, target demo $\mathcal{C}_{\mathrm{demo}}$, validation set $\mathcal{V}$
\Require Hidden full test set $\mathcal{T}$ \Comment{Not visible to builder}

\vspace{1.5mm}

\State Initialize builder workspace $\mathcal{W}_0\gets\{\mathcal{R},\mathcal{C}_{\mathrm{demo}},\mathcal{V}\}$
\State Initialize scaffold $S_0\gets\emptyset$ and index $k\gets 0$

\vspace{1.5mm}

\While{$M_{\mathrm{build}}$ scaffold is not submitted}
    \vspace{1mm}
    \Statex \quad\ \ \textbf{Step 1: Inspect task resources}
    \State $M_{\mathrm{build}}$ reads $\mathcal{R}$, $\mathcal{C}_{\mathrm{demo}}$, $\mathcal{V}$
    \Comment{Understand task}
    \vspace{1mm}
    \Statex \quad\ \ \textbf{Step 2: Propose or revise scaffold}
    \State $S_k \gets M_{\mathrm{build}}(\mathcal{W}_k)$
    \Comment{Implement scaffold}
    \vspace{1mm}
    \Statex \quad\ \ \textbf{Step 3: Evaluate on validation set}
    \State $\hat{Y}^{\mathcal{V}}_k \gets S_k(M_{\mathrm{tar}},\mathcal{V})$
    \State $a_k \gets \operatorname{Acc}(\hat{Y}^{\mathcal{V}}_k,Y^{\mathcal{V}})$
    \Comment{Call target model}
    \vspace{1mm}
    \Statex \quad\ \ \textbf{Step 4: Diagnose and improve}
    \State $\mathcal{E}_k \gets \{(x,y,\hat{y})\in\mathcal{V}: \hat{y}\neq y\}$
    \State $\mathcal{W}_{k+1}\gets \mathcal{W}_k\cup\{S_k,a_k,\mathcal{E}_k\}$
    \Comment{Refine scaffold}
    \vspace{1mm}
    \State $k\gets k+1$
\vspace{1mm}
\EndWhile

\vspace{1.5mm}

\Statex \textbf{Step 5: Export test-time entry point}
\State $\hat{S}\gets S_k$
\State Builder submits an executable entry point $f_{\hat{S}}(x;M_{\mathrm{tar}})$

\vspace{1.5mm}

\Statex \textbf{Step 6: Hidden evaluation}
\State Human evaluator runs $\hat{Y}^{\mathcal{T}}\gets f_{\hat{S}}(\mathcal{T};M_{\mathrm{tar}})$

\vspace{1.5mm}

\State \Return Test performance $\operatorname{Acc}(\hat{Y}^{\mathcal{T}},Y^{\mathcal{T}})$
\end{algorithmic}
\vspace{-1mm}
\noindent\rule{\linewidth}{0.6pt}

\vspace{-4mm}
\end{wraptable}

%% file: sections/setup.tex
\section{Experimental Setup}
\label{sec:setup}

\begin{figure*}[!t]
\centering
\vspace{0mm}
\begin{subfigure}{0.5\textwidth}
  \includegraphics[width=\textwidth]{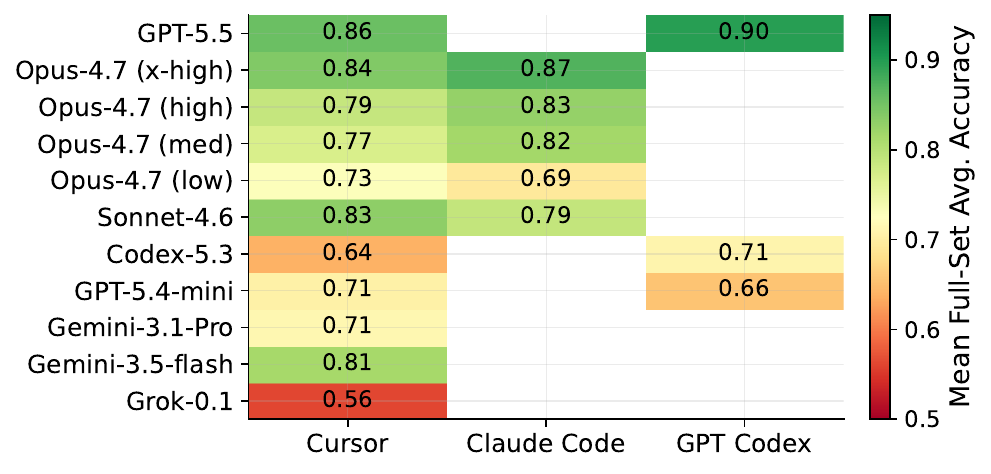}
  \vspace{-2mm}
  \caption{Builder $\times$ platform mean accuracy.}
\end{subfigure}
\hfill
\begin{subfigure}{0.48\textwidth}
  \vspace{0mm}
  \includegraphics[width=\textwidth]{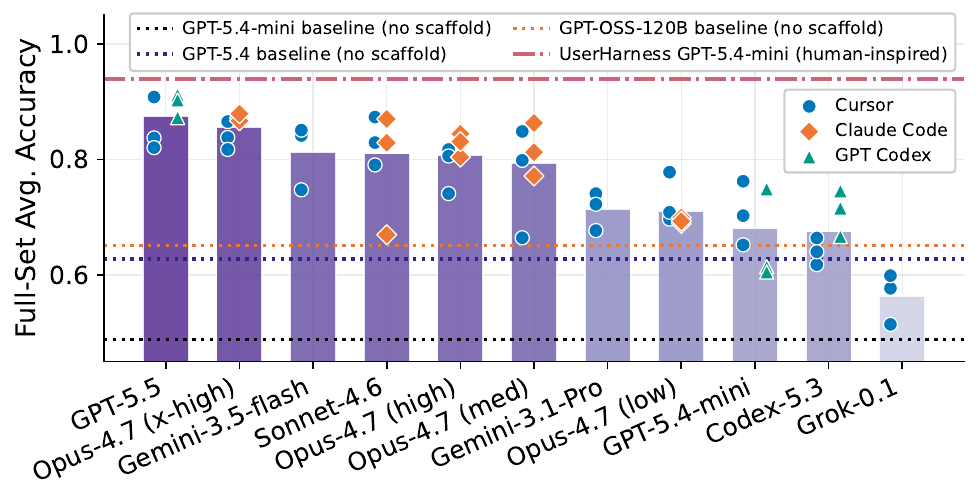}
  \vspace{-2mm}
  \caption{Builder ranking with per-run variation.}
\end{subfigure}
\vspace{-0mm}
\caption{Overview of the GPT-5.4-mini as target main results. (a) The mean accuracy of each evaluated builder--platform configuration, with blank cells indicating configurations that were not run. (b) The builder-level mean accuracy, with individual runs displayed as platform-colored markers. The dashed line denotes the reference baselines.}
\label{fig:overview}
\vspace{-0mm}
\end{figure*}

\textbf{Task and metric.}
The task aggregates four ToM datasets into a 3900-item hidden test set, including:
\begin{itemize}[topsep=-3pt, partopsep=-3pt, leftmargin=*, itemsep=-3pt]
\item \textbf{BigToM}~\citep{gandhi2023bigtom}: 1200 data points; binary belief/goal/action questions hinging on whether an agent observed a world change.
\item \textbf{Hi-ToM}~\citep{he2023hitom} 1200 data points; nested belief questions of recursion order $0$--$4$, with deception and multi-room object tracking.
\item \textbf{MMToM-QA}~\citep{jin2024mmtom}: 600 data points; binary Bayesian goal/belief inference from action trace.
\item \textbf{MuMA-Tom}~\citep{shi2024muma}: 900 data points; 3-choice multi-agent belief/social-goal/belief-of-goal questions.
\end{itemize}
We employ only the text format question of every task. Each builder additionally receives a $195$-item ($5\%$) validation sample drawn by a fixed random seed. The primary metric is the unweighted \emph{macro average} of the four per-benchmark full set accuracies; the number of validation evaluation uses is a secondary criterion (efficiency).

\textbf{Experiment design.}
We control the following hyper-parameters for each experiment run:
\begin{itemize}[topsep=-3pt, partopsep=-3pt, leftmargin=*, itemsep=-3pt]
  \item \textbf{Platform}: The harness that the builder model itself runs inside, including Cursor, Claude Code, and GPT Codex.
  \item \textbf{Builder Model}: The model that writes the scaffold for the targets, including Opus-4.7 (with different reasoning efforts from highest to lowest), Sonnet-4.6, GPT-5.5, GPT-5.4-mini, Codex-5.3, Gemini-3.1-Pro, Gemini-3.5-flash, and Grok-0.1. All the other models besides Opus-4.7 is experimented with highest reasoning effort.
  \item \textbf{Target Model}: The weaker model that is being scaffolded for performance improvement, including GPT-5.4-mini and Gemini-3.5-flash.
  \item \textbf{Repeats}: We repeat each experiment setting 3 times to investigate into the harness stability.
\end{itemize}
This yields in total 72 experiment runs. For the main setting, we use GPT-5.4-mini as the dominant target and serves as the common control for most controlled comparisons; Gemini-3.5-flash is used for the target-model only to contrast different target model's impact.

\textbf{Baselines.} We compare each scaffolded target model against two baseline settings:
\begin{itemize}[topsep=-3pt, partopsep=-3pt, leftmargin=*, itemsep=-3pt]
\item \textbf{Vanilla}: Each target model is called directly with the same naive prompt, without any task-specific scaffolding. This setting yields a macro-average accuracy of $\vbaseGptMacro{}$ for GPT-5.4-mini and $\vbaseGeminiMacro{}$ for Gemini-3.5-flash.
\item \textbf{Human-Inspired Harness}: Each target model is evaluated with UserHarness~\citep{qian2026userharness}, a human-designed harness framework for ToM problems. This setting yields a macro-average accuracy of $\vuserHarnessGptMacro{}$ for GPT-5.4-mini and $\vuserHarnessGeminiMacro{}$ for Gemini-3.5-flash.
\end{itemize}
The Vanilla baseline measures the performance that scaffolding is expected to improve upon, while the human-inspired harness provides a human-designed reference point for harness effectiveness.

%% file: sections/overview.tex
\section{Results and Analysis}
\label{sec:overview}

Before the detailed analyses, we first summarize the experimental design and main empirical patterns for the GPT-5.4-mini as the target model. \Cref{fig:overview}(a) reports the mean accuracy of each evaluated builder--platform configuration across three repeated runs. \Cref{fig:overview}(b) aggregates the same results by builder: each bar denotes the builder-level mean, while individual markers show the corresponding runs across platforms. We include four reference baselines for comparison: the no-scaffold baselines of GPT-5.4-mini, GPT-OSS-120B, and the stronger GPT-5.4 model, together with the human-inspired UserHarness scaffold applied to GPT-5.4-mini.

\begin{wraptable}{r}{0.45\linewidth}
\centering
\caption{Headline numbers of the main results, with target model GPT-5.4-mini.}
\label{tab:headline}
\setlength{\tabcolsep}{4pt}
\footnotesize
\begin{minipage}{\linewidth}
\input{tables/main_headline}
\end{minipage}
\end{wraptable}

Three observations are immediately visible and recur throughout the analysis. First, every evaluated builder--platform configuration substantially exceeds the GPT-5.4-mini no-scaffold baseline, showing that scaffolding consistently improves the weak target model. Second, the dominant source of variation is the builder model rather than the platform: builders form a clear vertical ordering, whereas platform-level differences within the same builder are comparatively small. Third, many scaffolded GPT-5.4-mini configurations surpass the no-scaffold GPT-5.4 baseline. This indicates that a well-designed scaffold can sometimes yield gains larger than upgrading to a stronger unscaffolded model, while the remaining gap to the human-inspired scaffold shows that automated scaffolding still has room for improvement.

%% file: tables/main_headline.tex
\resizebox{\linewidth}{!}{
\begin{tabular}{@{}p{0.66\linewidth}c@{}}
\toprule \textbf{Metric} & \textbf{Value} \\ \midrule
GPT-5.4-mini vanilla baseline & 0.488 \\
GPT-5.4 vanilla baseline & 0.619 \\
GPT-5.4-mini human-inspired & 0.939 \\ \midrule
Mean over all scaffolded runs & 0.763 (+0.275) \\ \midrule
Best run (GPT-5.5, GPT Codex) & 0.912 \\
-- uplift over baseline & +0.423 (86.7\%) \\ \midrule
Builder beating vanilla baseline & 100\% \\
\bottomrule \end{tabular}
}

%% file: sections/a0_main.tex
\subsection{Aspect 0: Main Results}
\label{sec:a0}

\begin{figure*}[!t]
\centering
\begin{minipage}[t]{0.53\linewidth}
    \centering
    \vspace{-0mm}
    \small
    \setlength{\tabcolsep}{3pt}
    \renewcommand{\arraystretch}{1.0}
    \resizebox{\linewidth}{!}{%
        \input{tables/main_by_builder}
    }
    \vspace{-0mm}
    \caption*{\textbf{(a)} Main results by builder model.}
\end{minipage}
\hfill
\begin{minipage}[t]{0.45\linewidth}
    \centering
    \vspace{-0mm}
    \includegraphics[width=\linewidth]{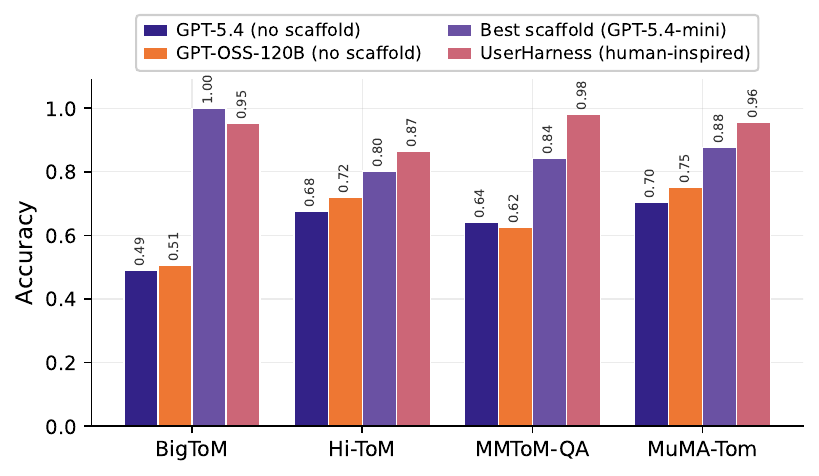}
    \vspace{-6mm}
    \caption*{\textbf{(b)} Per-benchmark comparison.}
\end{minipage}
\vspace{-0mm}
\caption{Main results by builder model and benchmark with GPT-5.4-mini as target.
(a) The performance by builder model, averaged over platforms$\times$repeats; $R$ denotes the number of runs pooled.
(b) The per-benchmark comparison between the best scaffold and three references: raw no-scaffold GPT-5.4, raw no-scaffold GPT-OSS-120B, and the human-inspired harness.}
\label{fig:main_results_combined}
\vspace{-0mm}
\end{figure*}

\setting{We fix the target model to GPT-5.4-mini, whose no-scaffold direct-call baseline has a macro-average accuracy of 0.488. For each builder model, we aggregate all the runs across platforms and repeats, and report both per-benchmark accuracy and macro-average accuracy. The no-scaffold direct-call result serves as the reference baseline.}

\result{
We present full results in \Cref{fig:main_results_combined}(a) and representative statistics in \Cref{tab:headline}. Strong-to-weak scaffolding yields a large and uniformly positive effect. Across all \vnRunsGpt{}\unskip\ scaffolded GPT-5.4-mini runs, the mean macro-average accuracy is $\vmeanScaffoldGpt{}$, corresponding to an uplift of $+\vmeanUpliftGpt{}$ over the baseline, and \textbf{\vfracBeatBase{}\ of runs exceed the baseline}. On average, every one of the \vnCells{}\ builder configurations improves over the baseline. The best individual run, produced by \vbestRunBuilder{}\ on \vbestRunPlatform{}, reaches $\vbestRunAvg{}$, an uplift of $+\vbestRunUplift{}$ ($\vbestRunRelUplift{}\%$ relative). Thus, scaffolding can lift a model that performs poorly on several tasks in the direct-call setting to near-ceiling performance.

\Cref{fig:main_results_combined}(b) reports the per-benchmark results against three reference points: the raw no-scaffold GPT-5.4 and GPT-OSS-120B baselines, and the human-designed UserHarness using the same GPT-5.4-mini backbone. The best automatically built scaffold outperforms both vanilla baselines on all four benchmarks, showing that scaffolding a smaller model can exceed the gains from simply moving to a much larger unscaffolded model. Relative to UserHarness, however, performance remains task-dependent. On BigToM, where the reasoning shortcut is explicit and can be consistently exploited, the automated scaffold reaches near-ceiling performance and slightly exceeds UserHarness ($1.00$ vs.\ $0.95$). Clear gaps remain on the more demanding benchmarks: Hi-ToM ($0.80$ vs.\ $0.87$), MMToM-QA ($0.84$ vs.\ $0.98$), and MuMA-ToM ($0.88$ vs.\ $0.96$). Thus, the remaining gap is concentrated in settings where the relevant reasoning is less amenable to compilation into deterministic structure, and where careful human harness engineering continues to provide an advantage.
}

\insight{The main result shows that strong-to-weak scaffolding yields a large and robust improvement. Since the builder never sees the hidden test set, the $+\vmeanUpliftGpt{}$ mean uplift indicates that the learned scaffold designs transfer beyond the validation slice. The strongest scaffold lifts GPT-5.4-mini above raw GPT-5.4 and GPT-OSS-120B on every benchmark, and matches the human-inspired harness on the structured BigToM task. This suggests that, when sub-problems are compilable, a builder's reasoning can be converted into reusable inference-time structure. Where such compilation is harder, however, the remaining gap to human-inspired harness persists.}

%% file: tables/main_by_builder.tex
\begin{tabular}{lccccccc}
\toprule
\textbf{Scaffold builder} & \textbf{$R$} & \textbf{BigToM} & \textbf{Hi-ToM} & \textbf{MMToM} & \textbf{MuMA} & \textbf{Avg.\ ($\pm$sd)} & \textbf{$\Delta$} \\
\midrule
\textit{Baseline (no scaffold)} & -- & 0.503 & 0.569 & 0.412 & 0.469 & 0.488 & -- \\
\midrule
GPT-5.5 & 6 & 1.000 & 0.803 & 0.842 & 0.857 & \textbf{0.875}\,$\pm$\,0.036 & +0.387 \\
Opus-4.7 (x-high) & 6 & 0.970 & 0.791 & 0.788 & 0.876 & \textbf{0.856}\,$\pm$\,0.022 & +0.368 \\
Gemini-3.5-flash & 3 & 0.986 & 0.712 & 0.778 & 0.777 & \textbf{0.813}\,$\pm$\,0.047 & +0.325 \\
Sonnet-4.6 & 6 & 0.977 & 0.712 & 0.742 & 0.810 & \textbf{0.810}\,$\pm$\,0.069 & +0.322 \\
Opus-4.7 (high) & 6 & 0.922 & 0.739 & 0.777 & 0.791 & \textbf{0.807}\,$\pm$\,0.033 & +0.319 \\
Opus-4.7 (med) & 6 & 0.944 & 0.699 & 0.751 & 0.778 & \textbf{0.793}\,$\pm$\,0.065 & +0.305 \\
Gemini-3.1-Pro & 3 & 0.910 & 0.732 & 0.618 & 0.593 & \textbf{0.713}\,$\pm$\,0.027 & +0.225 \\
Opus-4.7 (low) & 6 & 0.887 & 0.688 & 0.609 & 0.659 & \textbf{0.711}\,$\pm$\,0.031 & +0.222 \\
GPT-5.4-mini & 6 & 0.981 & 0.649 & 0.619 & 0.474 & \textbf{0.681}\,$\pm$\,0.062 & +0.193 \\
Codex-5.3 & 6 & 0.983 & 0.625 & 0.563 & 0.528 & \textbf{0.675}\,$\pm$\,0.043 & +0.187 \\
Grok-0.1 & 3 & 0.613 & 0.592 & 0.537 & 0.511 & \textbf{0.563}\,$\pm$\,0.036 & +0.075 \\
\bottomrule
\end{tabular}

%% file: sections/a1_stability.tex
\subsection{Aspect 1: Run-to-Run Stability}
\label{sec:a1}

\setting{We assess scaffold reproducibility by measuring the variance of the final full-set macro-average across independent repeats within the same setting, defined by the same platform, builder, and target model (GPT-5.4-mini).}

\begin{wrapfigure}{r}{0.52\linewidth}
\vspace{-2mm}
\centering

\includegraphics[width=\linewidth]{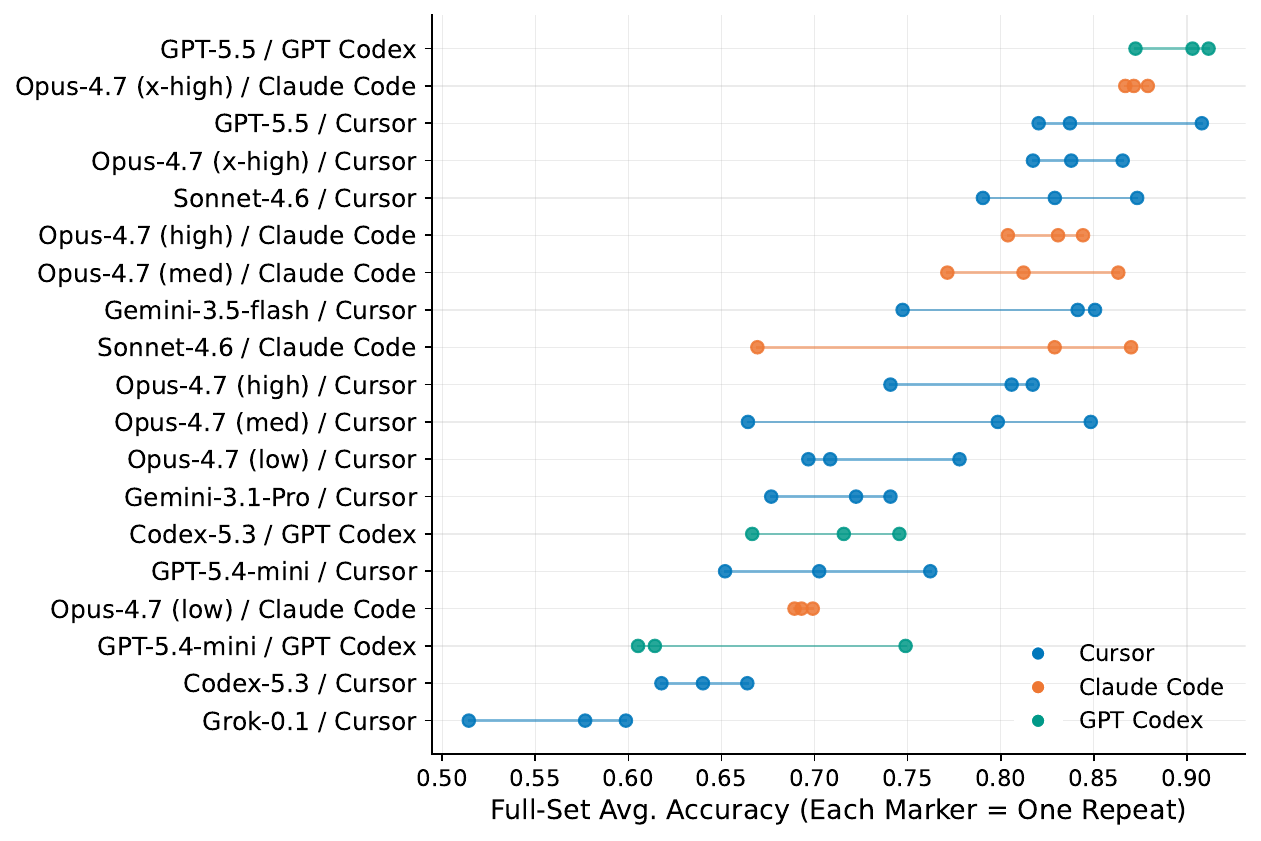}
\caption{Each marker shows one repeat's full-set macro-average for a
platform--builder combination setting, sorted by the average score;
tight clusters indicate reproducible builds, while long bars indicate
occasional weak scaffold builds.}
\label{fig:stability}

\vspace{0mm}

\captionof{table}{Refinement statistics by the builder model, including
validation runs, first/best validation accuracy, accuracy gain, and the
validation--full optimism gap (positive indicates validation
over-estimated the full set).}
\label{tab:refine}
\small
\vspace{1.5mm}
\input{tables/refine_by_builder}

\vspace{-8mm}
\end{wrapfigure}

\result{The standard deviation in \Cref{fig:main_results_combined}(a) and the visualization in \Cref{fig:stability} shows that the scaffold building procedure is fairly stable across platforms and builders. The mean standard deviation of all the macro-average is $\vmeanWithinSd{}$, roughly an order of magnitude smaller than the $+\vmeanUpliftGpt{}$ mean uplift. At the same time, the widest setting has a repeat range of $\vmaxWithinRange{}$, indicating that the build process is still not fully deterministic.}

\insight{Reproducibility is strong but not perfect, and the remaining instability is informative. Through case by case investigation, we discover that the largest spreads actually occur in settings where builders pursue \emph{deterministic-solver} strategies: a single logic error in a benchmark-specific rule can shift accuracy by tens of points over a 1000$+$ item benchmark full set. Prompt-only scaffolds are typically more stable, but they also deliver smaller gains. This points to a practical recipe: because failures are usually visible on validation and variance is moderate, building two or three scaffolds and selecting the best validation performer offers a low-cost way to capture the upper bound of a setting's performance range.}

%% file: tables/refine_by_builder.tex
\resizebox{\linewidth}{!}{
\begin{tabular}{lcccccc}
\toprule
\textbf{Builder} & \textbf{Val.\ Runs} & \textbf{First} & \textbf{Best} & \textbf{Gain} & \textbf{Val$-$Full Gap} \\
\midrule
Sonnet-4.6 & 6.8 & 0.521 & 0.856 & 0.334 & 0.045 \\
Gemini-3.1-Pro  & 6.7 & 0.374 & 0.712 & 0.338 & -0.002 \\
GPT-5.5  & 5.8 & 0.641 & 0.924 & 0.283 & 0.048 \\
Opus-4.7 (x-high)  & 5.5 & 0.668 & 0.888 & 0.221 & 0.032 \\
Opus-4.7 (med)  & 5.2 & 0.590 & 0.794 & 0.204 & 0.001 \\
GPT-5.4-mini  & 4.7 & 0.552 & 0.711 & 0.159 & 0.030 \\
Opus-4.7 (high)  & 4.5 & 0.615 & 0.842 & 0.227 & 0.035 \\
Codex-5.3  & 4.3 & 0.573 & 0.693 & 0.120 & 0.018 \\
Gemini-3.5-flash  & 4.0 & 0.545 & 0.845 & 0.300 & 0.032 \\
Opus-4.7 (low)  & 3.2 & 0.611 & 0.690 & 0.078 & -0.021 \\
Grok-0.1  & 2.7 & 0.344 & 0.556 & 0.212 & -0.007 \\
\bottomrule
\end{tabular}
}

%% file: sections/a2_refine.tex
\subsection{Aspect 2: Refinement on Validation}
\label{sec:a2}

\setting{We analyze the builder's record of its refinement trajectory on the validation set. Specifically, we capture how many validation evaluations the builder performed, how validation accuracy changed across iterations, and whether additional refinement translated into stronger final full-set performance. The target model is GPT-5.4-mini.}

\begin{figure*}[!t]
\centering
\begin{subfigure}{0.39\textwidth}
  \vspace{-0mm}
  \includegraphics[width=\textwidth]{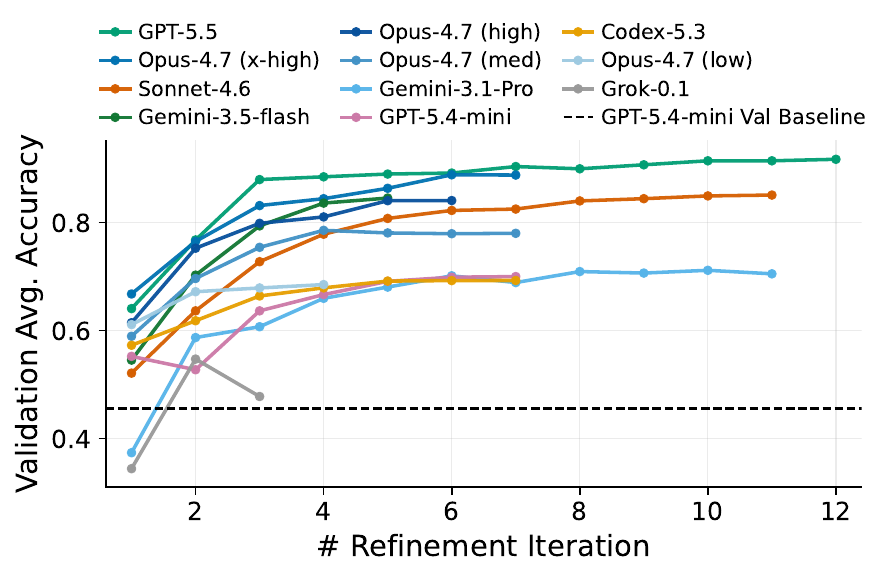}
  \caption{Mean validation trajectory per builder.}
\end{subfigure}
\hfill
\begin{subfigure}{0.59\textwidth}
  \vspace{-0mm}
  \includegraphics[width=\textwidth]{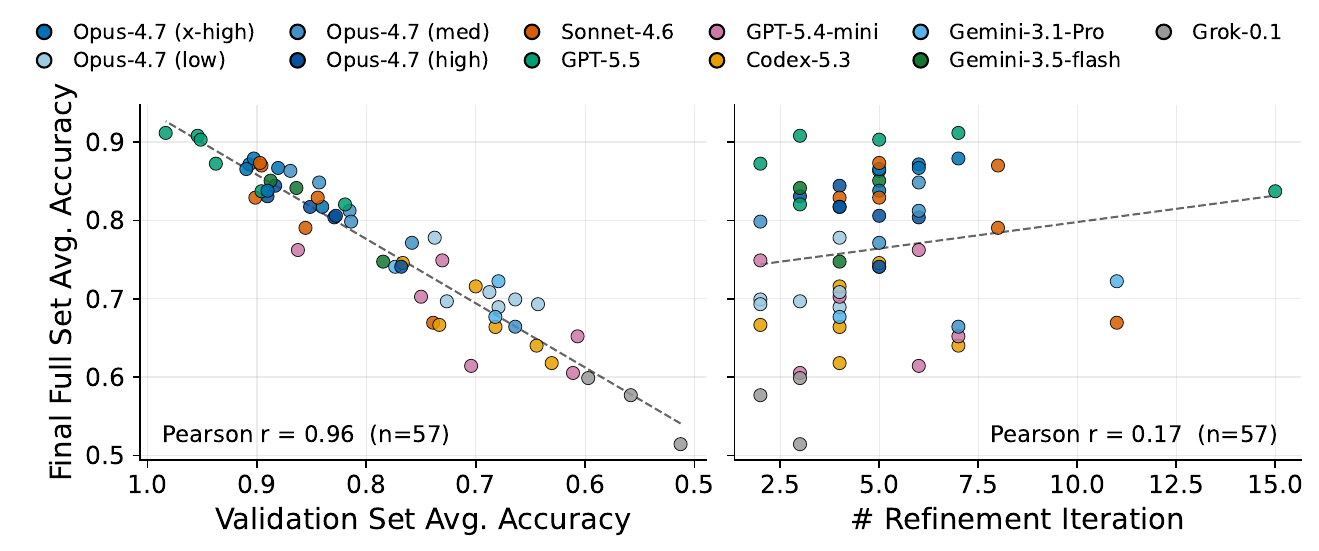}
  \caption{Validation accuracy and refinement budget vs.\ final full-set accuracy.}
\end{subfigure}
\vspace{-0mm}
\caption{Illustration of builder's refinement dynamics. (a) Validation accuracy increases over iterations for each builder, using carry-forward averages across repeats. (b) Two scatter plots share the same full-set accuracy axis: the best validation score closely tracks final full-set performance ($r=\vcorrValBestFull{}$), whereas the number of refinement iterations is largely unrelated to it ($r=\vcorrNvalFull{}$).}
\label{fig:refine_traj}
\label{fig:refine_scatter}
\vspace{-0mm}
\end{figure*}

\result{Builders use validation evaluations sparingly, as encouraged by the secondary scoring criterion: the mean number of validation passes is \vmeanNVal{} (median \vmedianNVal{}; range \vminNVal{}--\vmaxNVal{}). As shown in  Figure~\ref{fig:refine_traj}(a), within individual runs, refinement is productive: mean validation accuracy increases by $\vmeanValGain{}$ from the first logged iteration to the best logged iteration. Figure~\ref{fig:refine_scatter}(b) shows two complementary patterns against the same full-set performance axis. First, the best validation score is a strong proxy for held-out performance, tracking final full-set accuracy nearly one-to-one across runs (Pearson $r=\vcorrValBestFull{}$, left panel), with only a small optimism gap on average (mean $\vmeanValFullGap{}$). Thus, the 5\% validation sample effectively guides refinement without inducing substantial overfitting. Second, the amount of refinement itself is not predictive: the number of validation iterations is essentially uncorrelated with final full-set accuracy (Pearson $r=\vcorrNvalFull{}$, right panel). In short, builder quality matters much more than how often the builder probes the validation set.}

\insight{First, the near one-to-one relationship between best validation accuracy and final full-set accuracy, together with the small optimism gap, supports the study design: a frugal $5\%$ validation slice provides a faithful proxy for the hidden set, guiding refinement without substantial overfitting. Second, the flat relationship between validation budget and final performance shows that the limiting factor is not the amount of feedback, but the quality of the builder's hypotheses. Strong builders reach effective scaffolds through only a few principled refinements, whereas weaker builders do not reliably improve by probing the validation set more often. For scaffolding, therefore, the relevant form of test-time compute is not simply repeated validation querying, but the reasoning used to refine the scaffold itself.}

%% file: sections/a3_techniques.tex
\subsection{Aspect 3: Scaffolding Techniques}
\label{sec:a3}

\setting{We analyze every run by reading its scaffold code and optimization log, then coding the final scaffold using a fixed twelve-technique taxonomy based on our observation. This structured extraction covers all \vnRunsTotal{}\unskip\ main setting's runs. We report three levels of prevalence for scaffolds that uses GPT-5.4-mini as the target: the share of scaffolds using each technique, each builder's self-declared \emph{primary lever}, and the per-benchmark solution approach, categorized as deterministic, hybrid, model+rules, or model-only.}

\begin{wrapfigure}{r}{0.48\linewidth}
\centering
\small
\vspace{-2mm}
\begin{minipage}{\linewidth}
\centering
\label{fig:technique_prevalence}
\input{tables/technique_prevalence}
\vspace{-0mm}
\caption*{\textbf{(a)} Technique prevalence statistics across runs.}
\end{minipage}

\vspace{2mm}

\begin{minipage}{\linewidth}
\centering
\includegraphics[width=\linewidth]{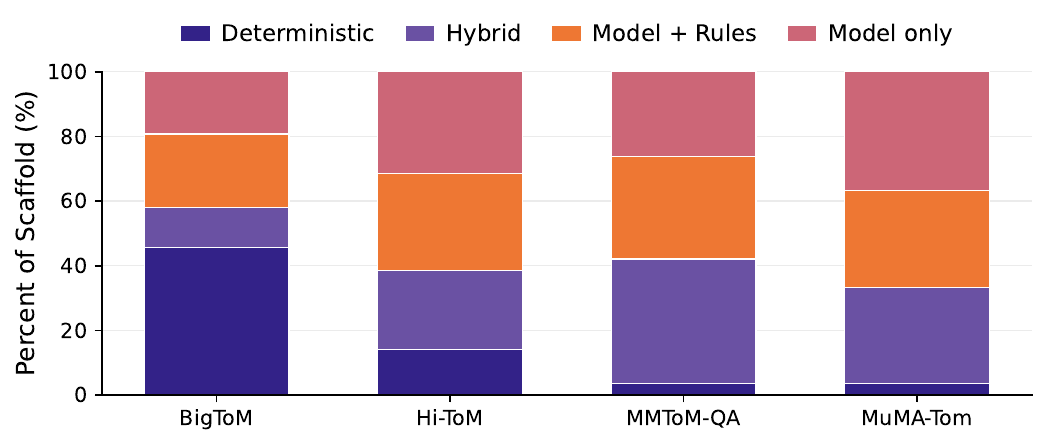}
\vspace{-4mm}
\caption*{\textbf{(b)} Solution approach across benchmarks.}
\label{fig:perbench_approach_panel}
\end{minipage}
\vspace{-1.5mm}
\caption{Technique prevalence and solution approach statistics across runs and benchmarks.}
\vspace{-12mm}
\label{fig:technique_and_approach}
\end{wrapfigure}

\result{From \Cref{fig:technique_and_approach}(a), we observe that two techniques are nearly universal: robust \emph{format enforcement}, which reliably parses the answer option, and \emph{greedy decoding}, implemented with temperature $0$. The next most common techniques are \emph{benchmark routing} and \emph{forced chain-of-thought}. More complex strategies are used less often: \emph{deterministic solvers}, \emph{self-consistency voting}, and \emph{verification/arbiter} passes appear only in a minority of scaffolds, while \emph{few-shot} prompting is also rare. The per-benchmark analysis in \Cref{fig:technique_and_approach}(b) shows that technique choice is strongly task-dependent. BigToM is often solved deterministically or with hybrid rules, since its question often exposes the observed/unobserved distinction, whereas MuMA-ToM is almost always handled by the model itself.}

\insight{The taxonomy suggests that effective scaffolding is less about exotic inference tricks than about disciplined task engineering. The most common techniques, such as format enforcement, greedy decoding, routing, and forced CoT, serve as inexpensive reliability controls: they prevent the weak target model from losing accuracy to malformed outputs, format confusion, or sampling variance. The techniques that most differentiate strong scaffolds require deeper task analysis. Deterministic solvers, structured state extraction, and polarity logic are useful only when the benchmark exposes enough regularity for the builder to convert reasoning into executable structure. In this sense, the per-benchmark approach map functions as a \emph{compilability ranking} of the ToM tasks: BigToM admits substantial rule-based compilation, whereas MuMA-ToM remains largely model-mediated. This distinction also explains the performance attribution results in later analysis.}

%% file: tables/technique_prevalence.tex
\resizebox{\linewidth}{!}{
\begin{tabular}{lcc}
\toprule
\textbf{Technique} & \textbf{Prevalence} & \textbf{Percent of Runs} \\
\midrule
Format enforcement & 57/57 & 100\% \\
Greedy / temp control & 56/57 & 98\% \\
Benchmark routing & 54/57 & 95\% \\
Forced CoT & 45/57 & 79\% \\
Polarity / negation logic & 45/57 & 79\% \\
Token-budget tuning & 43/57 & 75\% \\
Hybrid fallback & 34/57 & 60\% \\
Deterministic solver & 31/57 & 54\% \\
Structured extraction & 29/57 & 51\% \\
Few-shot examples & 12/57 & 21\% \\
Verification / arbiter & 7/57 & 12\% \\
Self-consistency vote & 3/57 & 5\% \\
\bottomrule
\end{tabular}
}

%% file: sections/a4_platform.tex
\subsection{Aspect 4: Harness Platform's Impact}
\label{sec:a4}

\begin{figure*}[t]
\centering
\begin{subfigure}{0.37\textwidth}
\vspace{-0mm}
\centering
\includegraphics[width=\linewidth]{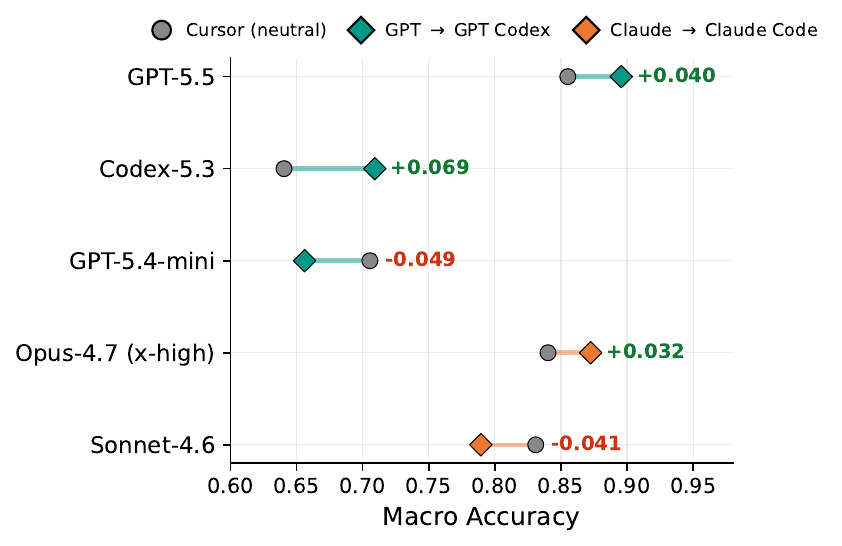}
\caption{Matched Native-vs-Cursor comparison.}
\label{fig:platform_native}
\end{subfigure}
\hfill
\begin{subfigure}{0.30\textwidth}
\vspace{-0mm}
\centering
\includegraphics[width=\linewidth]{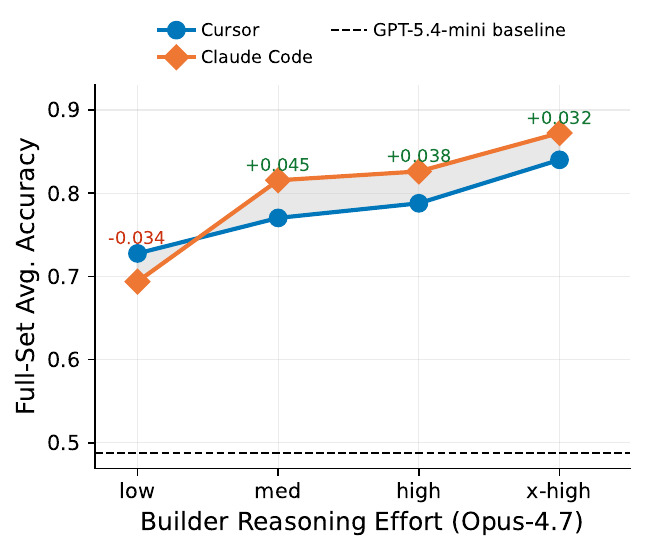}
\caption{Platform$\times$effort interaction.}
\label{fig:platform_effort}
\end{subfigure}
\hfill
\begin{subfigure}{0.31\textwidth}
\vspace{-0mm}
\centering
\includegraphics[width=\linewidth]{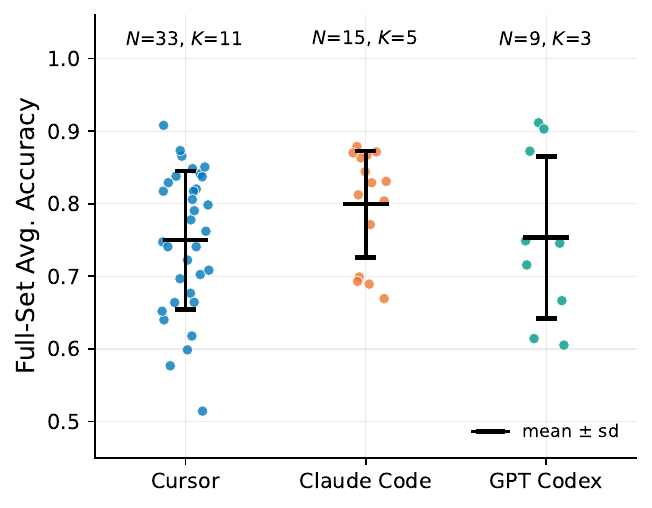}
\caption{Pooled platform distributions.}
\label{fig:platform_spread}
\end{subfigure}
\vspace{-0mm}
\caption{Platform effects with GPT-5.4-mini as the target. (a) Comparison of each builder's neutral Cursor run with its native-platform run, showing that matched native advantages are small and inconsistent. (b) We isolate Opus-4.7 across the full effort ladder, and shows the native-platform advantage emerges only at higher reasoning effort. (c) The pooled run distributions by platform. Overall, platform matters primarily as a second-order, conditional factor than as the main driver of scaffold quality.}
\vspace{-2mm}
\label{fig:platform}
\end{figure*}

\setting{The platform refers to the agentic coding environment in which the builder writes and refines the scaffold. Each builder family has a \emph{native} platform from the same vendor: GPT and Codex builders are native to GPT Codex, while Opus and Sonnet builders are native to Claude Code. Cursor serves as a neutral third-party platform shared across builders. Holding the target model fixed at GPT-5.4-mini, we ask whether builders perform better on their native platform or not. \Cref{fig:platform} evaluates this question from three perspectives: a matched native-vs-Cursor comparison for all builder runs, an Opus-4.7 model comparison across reasoning-efforts, and a pooled run distribution across platforms.}

\result{
(i) \textit{Native-platform gains are small and inconsistent.} In the matched comparison in \Cref{fig:platform}(a), each builder's Cursor run is paired with its native-platform run. The short connectors point in both directions, indicating no systematic native-platform advantage. Averaged across the eight matched configurations, moving from Cursor to the native platform changes macro accuracy by only $\vplatNativeAllDelta{}$, with the native platform winning in $\vplatNativeNBetter{}$ of $\vplatNativeNTotal{}$ cells (paired permutation test $p=\vplatNativePairedP{}$). The effect is somewhat larger for the GPT family ($\vplatNativeGptDelta{}$ on GPT Codex, driven mainly by Codex-5.3 at $+0.069$) than for the Claude family ($\vplatNativeClaudeDelta{}$ on Claude Code), but both effects are much smaller than the across-builder variation reported in the previous section. Moreover, each family also contains a counterexample: GPT-5.4-mini and Sonnet-4.6 both perform worse on their native platform after averaging parallel runs.

(ii) \textit{Platform advantages emerge only when the builder has enough reasoning budget to use them.} We show in \Cref{fig:platform}(b) the more informative pattern is a platform$\times$effort interaction rather than a uniform platform effect. For Opus-4.7, Claude Code trails Cursor at low effort ($\vplatEffortLowDelta{}$), but leads once the builder is allowed to deliberate more extensively: medium $\vplatEffortMedDelta{}$, high $\vplatEffortHighDelta{}$, and extra-high $\vplatEffortXhighDelta{}$. Thus, the native harness does not automatically improve the scaffold; its advantage materializes only when the builder can fully exploit its affordances.

(iii) \textit{Pooled platform averages mostly reflect builder composition.} The platform marginal in \Cref{fig:platform}(c) should therefore be interpreted descriptively rather than causally. The three platforms host different builder rosters ($K=11$, $5$, and $3$ builders), so their average scores conflate platform effects with builder selection. Claude Code has the highest marginal mean ($\vplatMargClaudecode{}$ vs.\ GPT Codex $\vplatMargGptcodex{}$ and Cursor $\vplatMargCursor{}$), but this mainly reflects its Opus/Sonnet-heavy roster rather than a clean platform advantage. The broad overlap among the platform clouds reinforces the main message: builder identity dominates platform identity.
}

\insight{The harness platform is a second-order factor, and its effect is conditional rather than universal. The matched comparisons provide little evidence for a general ``builders perform best on their own platform'' rule: the native advantage averages only $\vplatNativeAllDelta{}$, is not statistically reliable, and even reverses for some builders. Likewise, the apparent platform ranking in the pooled view mainly reflects roster composition rather than a causal platform effect. The more substantive finding is the platform$\times$effort interaction: a native harness helps only when the builder has enough reasoning budget to exploit its affordances. Practically, this makes the strong-to-weak recipe more portable than platform-specific explanations would suggest: the central determinants are still builder capability and reasoning effort, not the coding environment itself. At the same time, platform-specific tuning may still matter at the frontier, where a capable high-effort builder can convert better tooling into better scaffold design.}

%% file: sections/a5_weak.tex
\subsection{Aspect 5: Analysis Across Target Models}
\label{sec:a5}

\setting{We vary the target model from GPT-5.4-mini to include Gemini-3.5-flash, while holding the platform fixed to Cursor, so the only changing factor is \emph{which model is being scaffolded}. Five builders (Opus-4.7, GPT-5.5, Gemini-3.1-Pro, Gemini-3.5-flash, and Grok-0.1) were run against both targets, with three repeats each, yielding matched within-builder comparisons. The two targets begin from very different baselines: GPT-5.4-mini is weak overall ($\vbaseGptMacro{}$), whereas Gemini-3.5-flash is already strong, especially on Hi-ToM and MuMA-ToM ($\vbaseGeminiMacro{}$ overall). We therefore analyze not only the aggregate uplift, but also where the uplift occurs, how builders adapt their strategies, and when scaffolding can become harmful, using the different lenses we developed in previous analysis.}

\begin{figure*}[t]
\centering
\scriptsize
\captionsetup[subfigure]{justification=raggedright,singlelinecheck=false}

\begin{minipage}[t]{0.35\textwidth}
\vspace{-0mm}
\centering
\setlength{\tabcolsep}{1.5pt}
\small
\resizebox{\linewidth}{!}{%
\input{tables/weak_control}
}
\subcaption{Overall statistics of using the same builder on two target models, with $\Delta$ representing the macro uplift. }
\label{fig:weak_control}
\end{minipage}
\hfill
\begin{minipage}[t]{0.235\textwidth}
\vspace{-0mm}
\centering
\includegraphics[width=\linewidth]{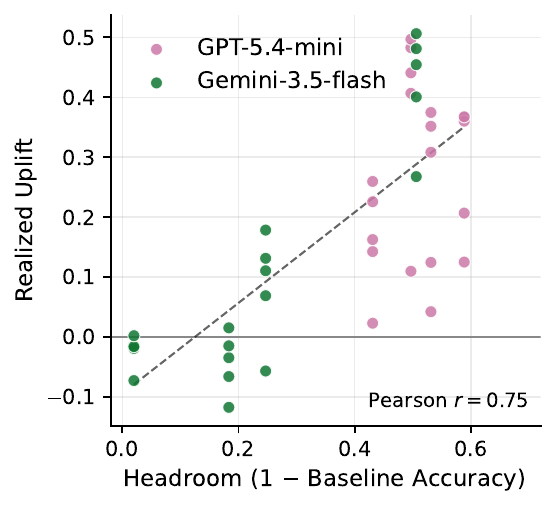}
\subcaption{The headroom law, with each point denotes one builder$\times$benchmark run.}
\label{fig:weak}
\end{minipage}
\hfill
\begin{minipage}[t]{0.38\textwidth}
\vspace{-0mm}
\centering
\setlength{\tabcolsep}{3pt}
\scriptsize
\resizebox{\linewidth}{!}{%
\input{tables/weak_perbench}
}
\centering
\subcaption{Per-benchmark accuracy uplift.}
\label{fig:weak_perbench}
\vspace{2mm}
\setlength{\tabcolsep}{3pt}
\small
\resizebox{\linewidth}{!}{%
\input{tables/weak_strategy}
}
\centering
\subcaption{Scaffold strategy sorted by targets.}
\label{fig:weak_strategy}
\end{minipage}

\vspace{-0mm}
\caption{
Weak-to-strong transfer analysis across target models. 
(a) Statistics of the same builder on two target models. GPT and Gemini denotes GPT-5.4-mini and Gemini-3.5-flash.
(b) The headroom law: each point is one builder$\times$benchmark run, with its realized uplift (scaffolded $-$ baseline accuracy) plotted against the headroom the target leaves on that benchmark ($1-\text{baseline}$).
(c) Per-benchmark statistics of the accuracy uplift. The weak target gains everywhere; the strong target gains essentially only on BigToM and even regresses on the tasks it already handles well.
(d) Builder's scaffold strategy sorted by different targets. Against the stronger Gemini, builders rely less on deterministic code and hand more benchmark tasks back to the model itself.
}
\label{fig:weak_all}
\vspace{-0mm}
\end{figure*}

\result{
As shown in \Cref{fig:weak_all}(a), overall, the weaker target benefits much more from scaffolding. Averaged over runs, scaffolding improves GPT-5.4-mini by $+\vweakGptUplift{}$ ($\vweakGptBase{}\rightarrow\vweakGptScaf{}$), but improves Gemini-3.5-flash by only $+\vweakGeminiUplift{}$ ($\vweakGeminiBase{}\rightarrow\vweakGeminiScaf{}$). This direction holds for all \vaFiveNBuilders{}\ builders. The aggregate contrast, however, is only the surface pattern; the per-benchmark results reveal a clearer mechanism.

(i) \textit{Uplift follows a headroom law.} As illustrated in \Cref{fig:weak_all}(b), across all builder$\times$benchmark$\times$target settings, realized uplift is strongly predicted by the target's available headroom on that benchmark, $1-\text{baseline}$ (Pearson $r=\vaFiveHeadroomCorr{}$). This suggests that scaffolding primarily recovers latent competence that the target model already possesses but does not reliably deploy, such as following the answer format, tracking observation cues, or maintaining recursive state. It therefore helps most where most correctable errors is left.

(ii) \textit{The location of the gain depends on the target.} In \Cref{fig:weak_all}(c) we show that, for GPT-5.4-mini, uplift is distributed across all four benchmarks. For Gemini-3.5-flash, however, the gain is concentrated almost entirely on BigToM, which accounts for \vaFiveGemBigtomShare\%\ of its macro uplift ($\vaFiveGemBigtomUplift{}$). This concentration is informative: BigToM is both the benchmark where the stronger target still has meaningful headroom and the task whose structure is most readily compiled into rules.

(iii) \textit{Builders adapt their strategy to the scaffolded target.} The coded taxonomy in \Cref{fig:weak_all}(d) shows that builders use less deterministic machinery when scaffolding Gemini-3.5-flash than when scaffolding GPT-5.4-mini. The share of model-only benchmark handling rises on every task, most sharply on MuMA-ToM ($40\%\rightarrow73\%$). In other words, builders appear to recognize that the stronger target can already solve Hi-ToM and MuMA-ToM benchmark tasks relatively well, and they reserve heavier rule-based interventions for the remaining compilable headroom.

(iv) \textit{On a strong target, scaffolding can backfire.} For GPT-5.4-mini, no builder regresses below baseline on any benchmark (\vaFiveGptRegCells/\vaFiveRegCellsTotal\ matched cases). For Gemini-3.5-flash, by contrast, every builder regresses on at least one benchmark (\vaFiveGemRegCells/\vaFiveRegCellsTotal\ cases), especially on tasks where the baseline is already high: Hi-ToM ($\vaFiveGemHitomUplift{}$ on average) and near-saturated MuMA-ToM ($\vaFiveGemMumaUplift{}$ on average). This illustrates the risk of over-scaffolding: when the target is already close to ceiling, additional prompts, routing, or rules may disrupt correct behavior more often than they repair errors.
}

\insight{The target model matters less through its identity than through its \emph{headroom}. Scaffolding acts primarily as a competence-recovery mechanism: its payoff is governed by how much latent ability the target fails to deploy, as reflected in the strong relationship between benchmark headroom and uplift. This turns the apparent rule that ``uplift shrinks as the target gets stronger'' into a more general principle: scaffolding helps when there are correctable failures left to recover. It also explains the observed adaptation in builder strategy. Capable builders allocate deterministic routing and rule-based machinery to sub-tasks where headroom remains, while backing off to lighter prompting where the target is already reliable. But when headroom is nearly exhausted, scaffolding can cross from assistance into interference, perturbing answers the model would otherwise get right. Practically, strong-to-weak scaffolding is therefore most valuable for weak targets; for already strong targets, it should be applied selectively at the sub-task level and gated by measured headroom.}

%% file: tables/weak_control.tex
\resizebox{\linewidth}{!}{
\begin{tabular}{llcccc}
\toprule
\textbf{Builder} & \textbf{Target} & \textbf{Baseline} & \textbf{Scaffolded} & \textbf{$\Delta$} \\
\midrule
Gemini-3.5-flash & GPT & 0.488 & 0.813 & +0.325 \\
Gemini-3.5-flash & Gemini & 0.761 & 0.872 & +0.111 \\
\midrule
Gemini-3.1-Pro & GPT & 0.488 & 0.713 & +0.225 \\
Gemini-3.1-Pro & Gemini & 0.761 & 0.881 & +0.120 \\
\midrule
GPT-5.5 & GPT & 0.488 & 0.855 & +0.367 \\
GPT-5.5 & Gemini & 0.761 & 0.901 & +0.140 \\
\midrule
Grok-0.1 & GPT & 0.488 & 0.563 & +0.075 \\
Grok-0.1 & Gemini & 0.761 & 0.780 & +0.019 \\
\midrule
Opus-4.7 (x-high) & GPT & 0.488 & 0.840 & +0.352 \\
Opus-4.7 (x-high) & Gemini & 0.761 & 0.923 & +0.162 \\
\bottomrule
\end{tabular}
}

%% file: tables/weak_perbench.tex
\resizebox{\linewidth}{!}{
\begin{tabular}{l ccc ccc}
\toprule
\multirow{2.5}{*}{\textbf{Benchmark}}  & \multicolumn{3}{c}{\textbf{GPT-5.4-mini}} & \multicolumn{3}{c}{\textbf{Gemini-3.5-flash}} \\
\cmidrule(lr){2-4}\cmidrule(lr){5-7}
& Base & Scaf. & $\Delta$ & Base & Scaf. & $\Delta$ \\
\midrule
BigToM & 0.50 & 0.89 & \better{+0.39} & 0.49 & 0.92 & \better{+0.42} \\
Hi-ToM & 0.57 & 0.73 & \better{+0.16} & 0.82 & 0.77 & \worse{-0.04} \\
MMToM-QA & 0.41 & 0.70 & \better{+0.28} & 0.75 & 0.84 & \better{+0.09} \\
MuMA-Tom & 0.47 & 0.71 & \better{+0.24} & 0.98 & 0.96 & \worse{-0.02} \\
\bottomrule
\end{tabular}
}

%% file: tables/weak_strategy.tex
\begin{tabular}{l c cccc}
\toprule
\multirow{2.5}{*}{\textbf{Target}} & \multirow{2.5}{*}{\textbf{\makecell{Det.-Solver\\Prevalence}}} & \multicolumn{4}{c}{\textbf{Model-only share by benchmark}} \\
\cmidrule(lr){3-6}
 & & BigToM & Hi-ToM & MMToM & MuMA \\
\midrule
GPT & 40\% & 27\% & 40\% & 27\% & 40\% \\
Gemini & 47\% & 40\% & 60\% & 53\% & 73\% \\
\bottomrule
\end{tabular}

%% file: sections/a6_effort.tex
\subsection{Aspect 6: Builder Reasoning Effort}
\label{sec:a6}

\setting{We fix the builder to Opus-4.7 and the target to GPT-5.4-mini, then vary only the builder's \emph{reasoning effort} across four tiers: low, medium, high, and extra-high. Because this sweep is available on both Cursor and Claude Code, it isolates how much the builder deliberates while writing the scaffold from both model identity and platform choice. Part of this comparison result is also shown in \Cref{fig:platform}(b).}

\begin{wraptable}{r}{0.52\linewidth}
\centering
\vspace{-2mm}
\caption{Detailed statistics of Opus-4.7 reasoning-effort sweep (with target GPT-5.4-mini) on both platforms, in ascending effort order. ``Val.\ Runs'' and ``Py LOC'' show how effort also changes the scaffold building process instead of just its score.}
\label{tab:effort}
\small
\vspace{-0mm}
\input{tables/effort_control}
\vspace{-0mm}
\end{wraptable}

\result{Greater builder reasoning effort consistently improves scaffold quality. As shown in \Cref{tab:effort}, macro accuracy increases monotonically on both platforms as effort rises. Pooled across platforms, performance moves from $\veffortLowMean{}$ at low effort to $\veffortMedMean{}$, $\veffortHighMean{}$, and $\veffortXhighMean{}$ at extra-high effort, yielding a strong monotone relationship between effort tier and per-run accuracy (Spearman $\rho=\veffortSpearman{}$). The extra-high tier significantly outperforms both the adjacent high tier ($\veffortXhighMean{}$ vs.\ $\veffortHighMean{}$; permutation test $p=\veffortXhighHighP{}$) and the low tier ($p=\veffortXhighLowP{}$), indicating that the trend is not driven by noise. The largest gain occurs from low to medium effort, while higher tiers provide smaller but still positive improvements. This suggests that most of the benefit comes from giving the builder enough deliberation to identify a viable scaffold strategy, with additional reasoning refining rather than transforming that strategy.}

\begin{figure*}[!t]
\centering
\small
\captionsetup[subfigure]{justification=centering,singlelinecheck=false}

\begin{minipage}[t]{0.48\textwidth}
\vspace{-0mm}
\centering
\small
\setlength{\tabcolsep}{3pt}
\renewcommand{\arraystretch}{0.95}
\resizebox{\linewidth}{!}{%
\input{tables/attribution}
}
\subcaption{Mean accuracy of runs using vs.\ not using each technique.}
\label{fig:attribution_table}
\end{minipage}
\hfill
\begin{minipage}[t]{0.51\textwidth}
\vspace{2mm}
\centering
\includegraphics[width=\linewidth]{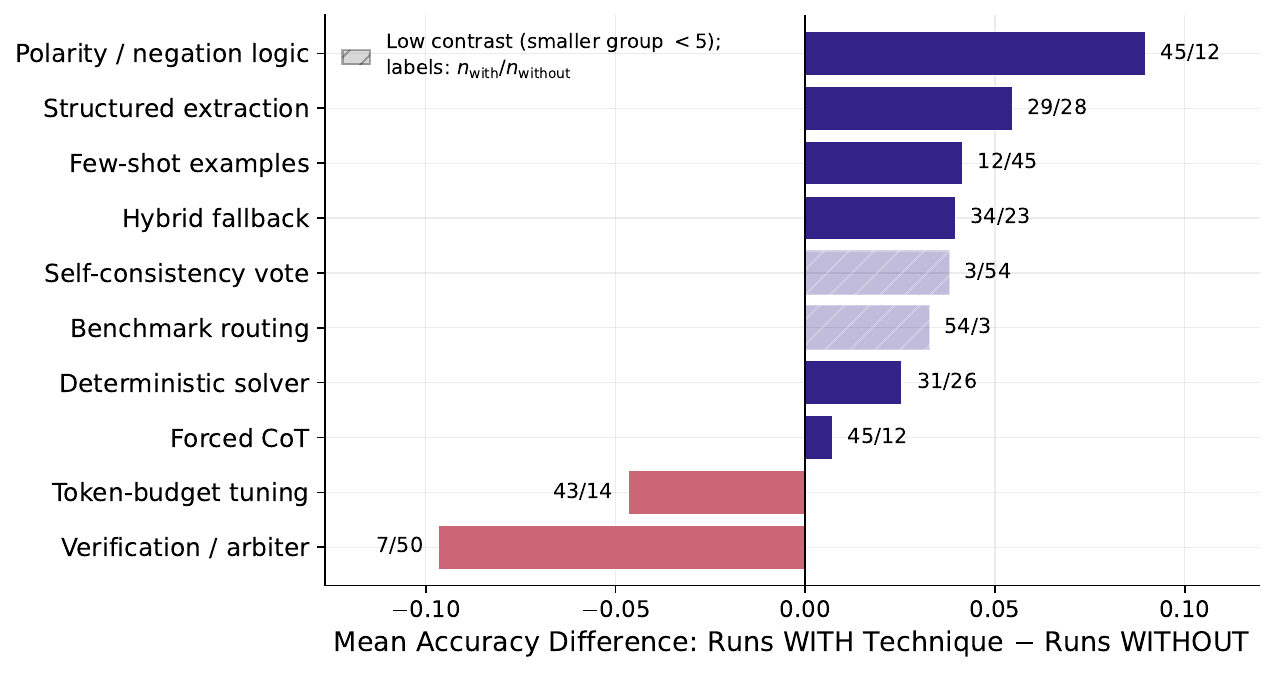}
\vspace{-1mm}
\subcaption{Association between each technique and accuracy.}
\label{fig:attribution_plot}
\end{minipage}

\vspace{-0mm}
\caption{Technique-level attribution analysis.
(a) Mean accuracy of runs using each technique vs.\ not using. $n_{\text{with}}$ denotes the number of runs using the technique. Associational/universal techniques have small $\Delta$ for lack of a contrast group.
(b) The technique's association with accuracy: difference in mean full-set accuracy between runs that use each technique and runs that do not.
}
\label{fig:attribution_all}
\vspace{-0mm}
\end{figure*}

\insight{Builder reasoning is a genuine monotonic lever: the more Opus-4.7 deliberates while \emph{writing} the scaffold, the stronger the resulting scaffold becomes, with no evidence of over-engineering even at extra-high effort. This contrasts with our previous analysis, which shows additional validation evaluations did not predict final quality. The useful compute is therefore not more probing, but deeper hypothesis formation about the task structure. The returns are diminishing but remain positive: the largest gain comes from low to medium effort, suggesting that moderate deliberation is enough to discover the main structural devices, such as routing, format enforcement, and deterministic solving, while higher tiers add more specialized refinements such as polarity logic and belief-state extraction. The accompanying growth in scaffold size ($\sim$510--650 LOC at low effort vs.\ $\sim$1000--1300 at extra-high) also supports this interpretation: greater builder reasoning compiles more task logic into the harness.}

%% file: tables/effort_control.tex
\resizebox{\linewidth}{!}{%
\begin{tabular}{llcccc}
\toprule
\textbf{Platform} & \textbf{Effort} & \textbf{$R$} & \textbf{Avg.\ ($\pm$sd)} & \textbf{Val.\ Runs} & \textbf{Py LOC} \\
\midrule
Cursor & low & 3 & 0.728\,$\pm$\,0.036 & 3.7 & 653 \\
Cursor & medium & 3 & 0.770\,$\pm$\,0.078 & 5.0 & 1037 \\
Cursor & high & 3 & 0.788\,$\pm$\,0.034 & 4.7 & 972 \\
Cursor & extra-high & 3 & 0.840\,$\pm$\,0.020 & 4.7 & 1274 \\
\midrule
Claude Code & low & 3 & 0.694\,$\pm$\,0.004 & 2.7 & 510 \\
Claude Code & medium & 3 & 0.816\,$\pm$\,0.038 & 5.3 & 685 \\
Claude Code & high & 3 & 0.826\,$\pm$\,0.017 & 4.3 & 870 \\
Claude Code & extra-high & 3 & 0.872\,$\pm$\,0.005 & 6.3 & 987 \\
\bottomrule
\end{tabular}
}

%% file: tables/attribution.tex
\begin{tabular}{lcccccc}
\toprule
\textbf{Technique} & \textbf{$n_{\text{w/}}$} & \textbf{$n_{\text{w/o}}$} & \textbf{Acc.\ w/} & \textbf{Acc.\ w/o} & \textbf{$\Delta$} & \textbf{Rel.} \\
\midrule
Polarity / negation logic & 45 & 12 & 0.782 & 0.693 & +0.090 &  \\
Structured extraction & 29 & 28 & 0.790 & 0.736 & +0.055 &  \\
Few-shot examples & 12 & 45 & 0.796 & 0.755 & +0.042 &  \\
Hybrid fallback & 34 & 23 & 0.779 & 0.740 & +0.040 &  \\
Self-consistency vote & 3 & 54 & 0.799 & 0.761 & +0.038 & $\dagger$ \\
Benchmark routing & 54 & 3 & 0.765 & 0.732 & +0.033 & $\dagger$ \\
Deterministic solver & 31 & 26 & 0.775 & 0.750 & +0.026 &  \\
Forced CoT & 45 & 12 & 0.765 & 0.758 & +0.007 &  \\
Token-budget tuning & 43 & 14 & 0.752 & 0.799 & -0.047 &  \\
Verification / arbiter & 7 & 50 & 0.679 & 0.775 & -0.097 &  \\
Greedy / temp control & 56 & 1 & 0.762 & 0.871 & -0.110 & $\dagger$ \\
Format enforcement & 57 & 0 & 0.763 & -- & -- & $\dagger$ \\
\bottomrule
\addlinespace
\multicolumn{7}{l}{\footnotesize $\dagger$ low contrast: smaller group $<5$ runs; $\Delta$ in this case is unreliable.}\\
\end{tabular}

%% file: sections/a7_attribution.tex
\subsection{Aspect 7: Attribution of Why Does Improvement Happen?}
\label{sec:a7}

\setting{We attribute uplift to scaffold techniques by comparing, for each taxonomy item, the mean full-set accuracy of GPT-5.4-mini-target runs that use the technique with those that do not. We also conduct three cross-checks: a paired McNemar test comparing the best scaffold to the no-scaffold baseline, a complementarity analysis measuring whether different scaffolds fix the same or different errors, and a comparison between self-scaffolding and stronger-builder scaffolding. Because techniques often co-occur, the technique-level comparisons are associational rather than strictly causal, but they provide a useful lens on which design choices align with higher performance.}

\begin{figure*}[!t]
\centering
\small
\captionsetup[subfigure]{justification=centering,singlelinecheck=false}

\begin{minipage}[t]{0.58\textwidth}
\vspace{-0mm}
\centering

\begin{minipage}[t]{\linewidth}
\vspace{0pt}
\centering
\setlength{\tabcolsep}{4pt}
\renewcommand{\arraystretch}{1.15}
\small
\resizebox{\linewidth}{!}{%
\input{tables/significance}
}
\subcaption{Statistical significance of the best scaffold compared with the baseline.}
\label{fig:significance}
\end{minipage}

\vspace{3mm}

\begin{minipage}[t]{\linewidth}
\vspace{0pt}
\centering
\setlength{\tabcolsep}{2.5pt}
\renewcommand{\arraystretch}{1.15}
\small
\resizebox{\linewidth}{!}{%
\input{tables/self_vs_strong}
}
\subcaption{Comparison between self-scaffolding and stronger builders.}
\label{fig:self_vs_strong}
\end{minipage}

\end{minipage}
\hfill
\begin{minipage}[t]{0.41\textwidth}
\vspace{0mm}
\centering
\includegraphics[width=0.92\linewidth]{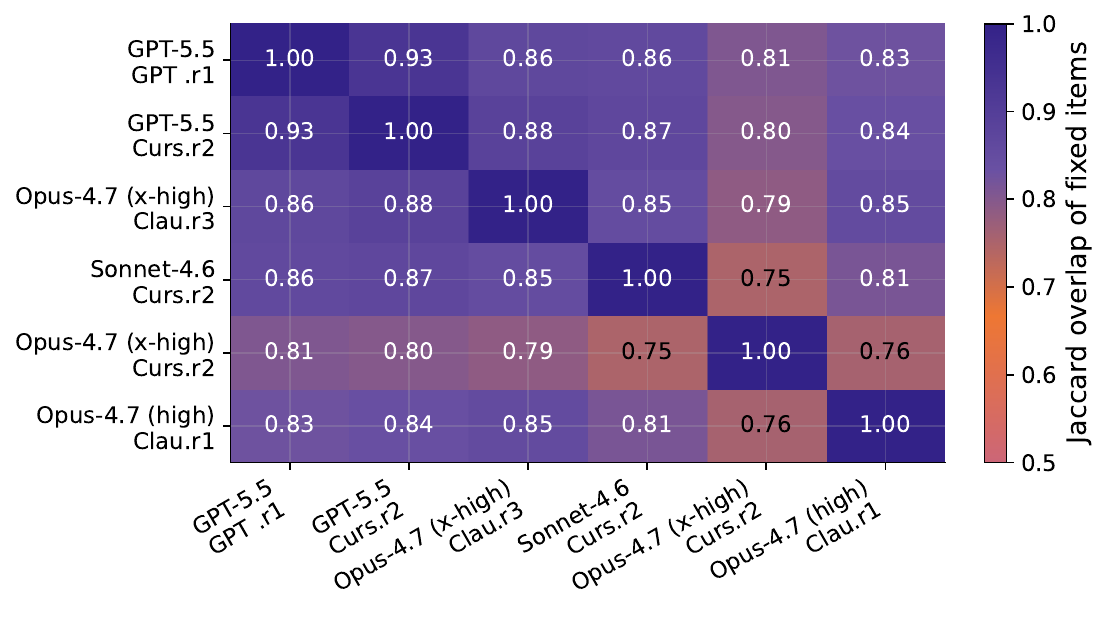}
\vspace{-0mm}
\subcaption{Complementarity among repaired errors.}
\label{fig:complementarity}
\end{minipage}

\vspace{-0.0em}
\caption{
Complementarity and builder-strength analyses. 
(a) Statistical significance of the best scaffold compared with the baseline, measured by paired McNemar tests over 3{,}900 items; ``Fixed'' denotes baseline-wrong cases corrected by the scaffold, while ``Broke'' denotes baseline-right cases made incorrect. 
(b) Comparison between self-scaffolding, where GPT-5.4-mini builds for itself, and stronger builders under the same platform and target setting; $\Delta$ reports the uplift over GPT-5.4-mini's own baseline. 
(c) Pairwise Jaccard overlap among the baseline-error sets repaired by top scaffolds; although scaffolds share many easy fixes, their union covers more baseline errors than any individual scaffold, indicating complementary repair mechanisms.
}
\label{fig:significance_self_complementarity}
\vspace{-1mm}
\end{figure*}

\result{(i) \textit{The strongest associations come from techniques that compile task structure into the scaffold.} \Cref{fig:attribution_all}(a) and (b) both compare runs with and without each technique, marking low-contrast comparisons whose smaller group has fewer than five runs. Among the better-powered contrasts, the largest positive associations are \emph{polarity/negation logic} ($+0.09$), \emph{structured extraction} ($+0.06$), \emph{hybrid fallback} ($+0.04$), and \emph{deterministic solving}. These are not generic prompting tricks; they directly target the main failure modes of the ToM benchmarks, including MOST-vs-LEAST framing, belief-state tracking, and offloading predictable sub-problems into code. Apparent negative associations for near-universal techniques such as greedy decoding ($56/1$) and benchmark routing ($54/3$) should not be interpreted as harmful effects: their contrast groups are too small and consist of unusually weak non-user runs.

(ii) \textit{The best scaffold produces a large, item-level reliable improvement.} As shown in \Cref{fig:significance_self_complementarity}(a), the gain over the GPT-5.4-mini no-scaffold baseline is overwhelmingly significant under a paired McNemar test over the 3900 evaluation items, with $\chi^2 \gg 10^4$ and $p < 10^{-4}$. The direction of the item-level changes is also highly asymmetric: the scaffold fixes \vmcnemarGptFixed{}\ baseline errors while breaking only \vmcnemarGptBroke{}\ previously correct items. Thus, the aggregate uplift is not driven by noise or by a small error redistribution, but reflects a broad shift from incorrect to correct predictions.

(iii) \textit{A stronger builder is not necessary for uplift, but it is necessary for the highest gains.} The self-scaffold control in \Cref{fig:significance_self_complementarity}(b), where GPT-5.4-mini builds a scaffold for itself, already improves performance by $+0.17$ to $+0.22$ over the no-scaffold baseline. This shows that even a weak target can use validation feedback and task structure to engineer a useful harness. However, stronger builders achieve substantially larger gains on both platforms, with the largest gap on GPT Codex ($+0.31$ for stronger builders vs.\ $+0.17$ for self-scaffolding). Self-scaffolding therefore recovers some accessible structure, but stronger builders are what unlock the high-performance regime.

(iv) \textit{Different strong scaffolds capture partly different ToM skills.} The complementarity analysis in \Cref{fig:significance_self_complementarity}(c) shows that top scaffolds overlap on many easy fixes but diverge on harder cases. The union of items fixed across the top scaffolds covers \voracleFixCov{}\ of all baseline errors, exceeding the coverage of any individual scaffold. This indicates that scaffold designs are not merely redundant variants of the same solution: different builders discover partially distinct mechanisms for repairing the target model's reasoning.
}

\insight{The gains come from two complementary layers of scaffold design. The first is a reliability floor: format enforcement, routing, and greedy decoding are used by nearly all scaffolds and prevent avoidable errors from malformed outputs, format confusion, or sampling variance. Because these techniques are almost universal, they do not explain variation across runs, but they make higher performance possible. The second layer is task-structure exploitation: polarity logic, structured belief-state extraction, and deterministic solving distinguish the strongest scaffolds by converting benchmark regularities into explicit inference-time machinery. This interpretation is reinforced by the significance and complementarity analyses. The best scaffold produces a large and reliable item-level improvement, yet different strong scaffolds repair overlapping but non-identical subsets of baseline errors. Thus, the ceiling is not determined by any single technique, but by the residual cases that remain difficult across diverse scaffold designs.}

%% file: tables/significance.tex
\begin{tabular}{llcccccc}
\toprule
\textbf{Target} & \textbf{Best Scaffold} & \textbf{Base} & \textbf{Scaf.} & \textbf{Fixed} & \textbf{Broke} & \textbf{$\chi^2$} & \textbf{$p$} \\
\midrule
GPT-5.4-mini & GPT-5.5/GPT Codex & 0.488 & 0.912 & 1717 & 105 & 1424.4 & $<10^{-4}$ \\
Gemini-3.5-flash & Gemini-3.5-flash/Cursor & 0.761 & 0.939 & 772 & 63 & 600.3 & $<10^{-4}$ \\
\bottomrule
\end{tabular}

%% file: tables/self_vs_strong.tex
\begin{tabular}{llccc @{\hspace{2mm}}|@{\hspace{2mm}} llccc}
\toprule
\textbf{Platform} & \textbf{Builder} & \textbf{$R$} & \textbf{Avg.\ ($\pm$sd)} & \textbf{$\Delta$} &
\textbf{Platform} & \textbf{Builder} & \textbf{$R$} & \textbf{Avg.\ ($\pm$sd)} & \textbf{$\Delta$} \\
\midrule
Cursor & Self & 3 & 0.706\,$\pm$\,0.045 & +0.217 & Codex & Self & 3 & 0.656\,$\pm$\,0.066 & +0.168 \\
& Others & 30 & 0.754\,$\pm$\,0.098 & +0.266 & & Others & 6 & 0.802\,$\pm$\,0.097 & +0.314 \\
\bottomrule
\end{tabular}

%% file: sections/a8_cogload.tex
\subsection{Aspect 8: Cognitive-Load Reduction}
\label{sec:a8}

\setting{In the task instructions, builders are explicitly asked to reduce the target model's cognitive load. We operationalize this idea using scaffold provenance tags. After evaluation, every runis labeled with a method that produced it, such as ``deterministic'', ``model\_prompt'', ``rule'', or ``llm''. For all these runs, we compute the determinism fraction: the share of the 3900 evaluation items answered entirely by code or structured rules. We then relate this to final accuracy and examine how it varies by benchmark. The target is GPT-5.4-mini for all runs.}

\begin{wraptable}{r}{0.48\linewidth}
\centering
\vspace{-4mm}
\caption{Statistics about determinism fraction, final accuracy, and scaffold code size for runs with prediction-provenance tags. Determinism fraction is the share of items answered through deterministic codes and rules.}
\label{tab:determinism}
\setlength{\tabcolsep}{3.5pt}
\renewcommand{\arraystretch}{1.0}
\small
\input{tables/determinism}
\vspace{-0mm}
\end{wraptable}

\begin{figure*}[!t]
\centering
\vspace{-0mm}
\begin{subfigure}{0.49\textwidth}
  \includegraphics[width=\textwidth]{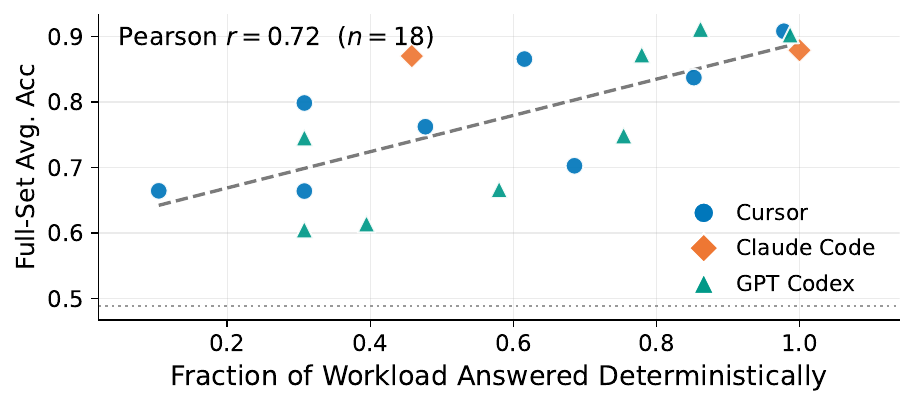}
  \caption{Determinism fraction vs.\ accuracy.}
  \label{fig:determinism_acc_scatter}
\end{subfigure}\hfill
\begin{subfigure}{0.49\textwidth}
  \includegraphics[width=\textwidth]{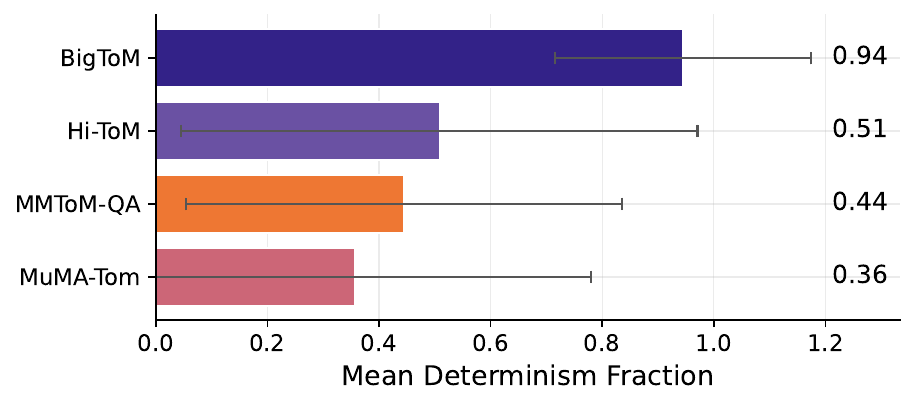}
  \caption{Benchmark-level offloadability.}
  \label{fig:det_by_bench}
\end{subfigure}
\caption{Deterministic offloading as a measure of cognitive-load reduction.
(a) scaffolds answering a larger share of items without target-model calls tend to achieve higher final accuracy.
(b) offloadability varies sharply by benchmark: BigToM is almost fully compilable into code, whereas MuMA-ToM remains substantially model-dependent.
}
\label{fig:determinism_acc}
\vspace{-0mm}
\end{figure*}

\result{Deterministic offloading is strongly associated with scaffold quality. As shown in Figure~\ref{fig:determinism_acc}(a), runs with higher determinism fractions achieve higher final accuracy (Pearson $r=0.72$): the more work the builder converts into executable structure, the less reasoning burden remains for the weak target model. According to Figure~\ref{fig:determinism_acc}(b), this relationship is also highly benchmark-dependent. BigToM is almost fully offloadable, with mean determinism around $0.94$, because its questions often expose the observed/unobserved distinction needed for many answers. Hi-ToM is partially offloadable ($\approx 0.51$), typically through symbolic belief-state tracking. MMToM-QA is similar but slightly lower ($\approx 0.44$), while MuMA-ToM is least reducible to structured code ($\approx 0.36$), reflecting the difficulty of compiling free-form dialogue reasoning into deterministic rules. Statistics in Table~\ref{tab:determinism} further shows that scaffold code size is only weakly related to accuracy ($r\approx 0.22$). Thus, what matters is not simply writing more code, but writing code that removes the right cognitive load from the target model.}

\insight{This is the mechanistic core of strong-to-weak scaffolding. The builder reduces the weak target model's cognitive load in two ways. The first is \emph{offloading}: deterministic codes and rules answer some items directly without target model's additional reasoning efforts. The second is \emph{structuring}: when the model is still needed, the scaffold narrows the task into a more constrained prompt with clearer inputs, output format, and reasoning focus. Offloading explains much of the variation in scaffold quality ($r=0.72$), but its feasibility depends on how much of the task can be compiled into explicit structure. In effect, the builder pays a one-time reasoning cost to encode part of the task's decision procedure into the harness; after that, the per-item burden on the weak model is reduced, and the model is reserved for the cases that resist rule or skill compilation.}

%% file: tables/determinism.tex
\resizebox{\linewidth}{!}{
\begin{tabular}{llccc}
\toprule
\textbf{Builder} & \textbf{Platform} & \textbf{Det.\ Frac.} & \textbf{Acc.} & \textbf{Py LOC} \\
\midrule
Opus-4.7 (x-high) & Claude Code & 1.00 & 0.879 & 1052 \\
GPT-5.5 & GPT Codex & 0.99 & 0.903 & 1285 \\
GPT-5.5 & Cursor & 0.98 & 0.908 & 1026 \\
GPT-5.5 & GPT Codex & 0.86 & 0.912 & 1288 \\
GPT-5.5 & Cursor & 0.85 & 0.837 & 1107 \\
GPT-5.5 & GPT Codex & 0.78 & 0.872 & 1294 \\
GPT-5.4-mini & GPT Codex & 0.75 & 0.749 & 1170 \\
GPT-5.4-mini & Cursor & 0.69 & 0.702 & 1462 \\
Opus-4.7 (x-high) & Cursor & 0.62 & 0.865 & 1293 \\
Codex-5.3 & GPT Codex & 0.58 & 0.667 & 1028 \\
GPT-5.4-mini & Cursor & 0.48 & 0.762 & 1596 \\
Sonnet-4.6 & Claude Code & 0.46 & 0.870 & 856 \\
GPT-5.4-mini & GPT Codex & 0.39 & 0.614 & 1129 \\
Codex-5.3 & Cursor & 0.31 & 0.664 & 1026 \\
Codex-5.3 & GPT Codex & 0.31 & 0.746 & 903 \\
GPT-5.4-mini & GPT Codex & 0.31 & 0.605 & 779 \\
Opus-4.7 (med) & Cursor & 0.31 & 0.798 & 796 \\
Opus-4.7 (med) & Cursor & 0.10 & 0.664 & 1553 \\
\bottomrule
\end{tabular}
}

%% file: sections/a9_errors.tex
\subsection{Aspect 9: Remaining Error Analysis}
\label{sec:a9}

\setting{We analyze the residual errors of the \vnTopScaffolds{}\ strongest GPT-5.4-mini scaffolds, defined as the best repeat from each platform$\times$builder setting with mean accuracy above $0.80$. We pool their remaining errors and slice them into multiple subcategories: BigToM by question type and observed/unobserved status, Hi-ToM by recursion order and deception, MMToM-QA by question subtype, and MuMA-ToM by label. We also decompose each scaffold's impact into baseline errors \emph{fixed} and baseline-correct items \emph{broken} to distinguish genuine improvement from error trade-offs.}

\begin{wraptable}{r}{0.45\linewidth}
\centering
\vspace{-2mm}
\caption{Residual accuracy by fine-grained metadata slice, pooled over the \vnTopScaffolds{}\ strongest scaffolds. Slices are grouped by benchmark, and $n$ denotes the pooled item count.}
\label{tab:error_slices}
\setlength{\tabcolsep}{3pt}
\renewcommand{\arraystretch}{1.0}
\small
\input{tables/error_slices}
\vspace{-2mm}
\end{wraptable}

\begin{figure*}[t]
\centering
\vspace{-0mm}
\begin{subfigure}{0.48\textwidth}
  \includegraphics[width=\textwidth]{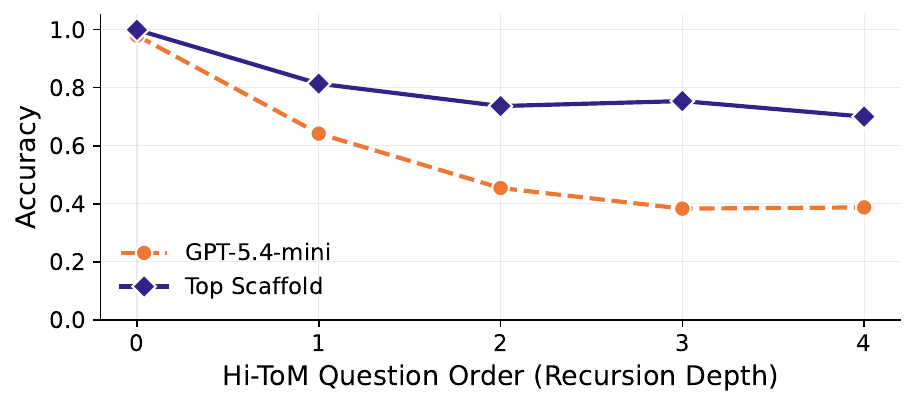}
  \caption{Hi-ToM recursion depth.}
  \label{fig:hitom_order_panel}
\end{subfigure}\hfill
\begin{subfigure}{0.47\textwidth}
  \includegraphics[width=\textwidth]{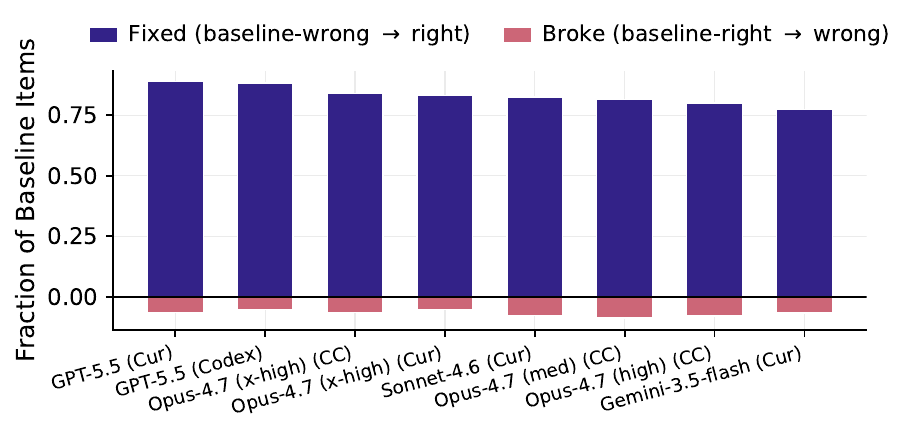}
  \caption{Fixes versus regressions.}
  \label{fig:fix_break_panel}
\end{subfigure}
\vspace{-0mm}
\caption{Residual-error structure among the strongest GPT-5.4-mini scaffolds.
(a) Hi-ToM accuracy declines as recursion order increases, and that the scaffold advantage narrows on the deepest cases.
(b) Results of decomposing each top scaffold's effect into baseline-wrong items fixed and baseline-correct items broken, showing that top scaffolds repair many more errors than introduce.
}
\label{fig:hitom_order}
\label{fig:fix_break}
\vspace{-0mm}
\end{figure*}

\result{
(i) \textit{Top scaffolds are broadly corrective rather than merely redistributive.} As shown in Figure~\ref{fig:fix_break}(b), the strongest scaffolds fix a large fraction of baseline mistakes while introducing few regressions: on average, they repair \vmeanFracFixed{}\ of baseline-wrong items and break only \vmeanFracBroke{}\ of baseline-correct items. This asymmetry indicates that scaffolding is close to Pareto-improving over the baseline rather than simply shifting errors across examples.

(ii) \textit{The remaining errors are concentrated in the least compilable parts of the tasks.} Table~\ref{tab:error_slices} shows that BigToM is essentially solved, with accuracy at least $0.95$ on every slice. The residual error mass instead clusters in three harder regions. First, as shown in Figure~\ref{fig:hitom_order}(a), Hi-ToM accuracy declines with recursion depth, falling from $\vhitomOrderZero{}$ at order 0 to $\vhitomOrderFour{}$ at order 4, with deception further reducing performance. Second, MMToM-QA errors concentrate in Bayesian goal-inference subtypes, especially the type-2 ``which container/goal'' questions. Third, MuMA-ToM remains difficult on social-goal and belief-of-goal labels, even though simpler belief questions are nearly solved.
}

\insight{The residual error floor marks the boundary of what current scaffolds can compile away. Scaffolding succeeds when the builder can turn recurring structure into deterministic rules, skills, routing logic, or constrained prompts. It struggles when the task requires nested higher-order belief tracking under deception or Bayesian goal inference from ambiguous action traces. In these cases, the scaffold must leave more of the reasoning to GPT-5.4-mini, precisely where the weak target remains least reliable. The strong fix/break asymmetry shows that the scaffolds are genuinely improving the target rather than trading one class of mistakes for another. At the same time, the complementarity result from previous analysis, where the union of top-scaffold fixes covers \voracleFixCov{}\ of baseline errors, suggests that some remaining failures are addressable by combining diverse scaffolds. The deepest recursion and goal-inference errors, however, likely require a stronger explicit belief-tracking mechanism than more variants of the same scaffold design.}

%% file: tables/error_slices.tex
\resizebox{\linewidth}{!}{
\begin{tabular}{llcc}
\toprule
\textbf{Benchmark} & \textbf{Slice} & \textbf{Accuracy} & \textbf{$n$ (pooled)} \\
\midrule
BigToM & goal/observed & 0.955 & 1600 \\
BigToM & belief/observed & 0.980 & 1600 \\
BigToM & action/observed & 0.988 & 1600 \\
BigToM & goal/unobserved & 0.993 & 1600 \\
BigToM & belief/unobserved & 0.996 & 1600 \\
BigToM & action/unobserved & 0.998 & 1600 \\
\midrule
Hi-ToM & order 0 & 0.999 & 1920 \\
Hi-ToM & order 1 & 0.814 & 1920 \\
Hi-ToM & order 2 & 0.736 & 1920 \\
Hi-ToM & order 3 & 0.754 & 1920 \\
Hi-ToM & order 4 & 0.700 & 1920 \\
Hi-ToM & deception=False & 0.829 & 4800 \\
Hi-ToM & deception=True & 0.772 & 4800 \\
\midrule
MMToM & qtype 2.1 & 0.680 & 600 \\
MMToM & qtype 2.4 & 0.755 & 600 \\
MMToM & qtype 1.3 & 0.790 & 800 \\
MMToM & qtype 1.2 & 0.800 & 800 \\
MMToM & qtype 2.3 & 0.828 & 600 \\
MMToM & qtype 2.2 & 0.863 & 600 \\
MMToM & qtype 1.1 & 0.929 & 800 \\
\midrule
MuMA & social\_goal & 0.872 & 1616 \\
MuMA & belief\_of\_goal & 0.880 & 3968 \\
MuMA & belief & 0.985 & 1616 \\
\bottomrule
\end{tabular}
}

%% file: sections/conclusion.tex
\section{Synthesis and Conclusion}
\label{sec:conclusion}

Across \vnRunsTotal{}\unskip\ runs, the evidence for \textbf{strong-to-weak scaffolding} is consistent and mechanistically interpretable. The main conclusions are as follows:

\begin{itemize}[topsep=-3pt, partopsep=-3pt, leftmargin=*, itemsep=-3pt]
  \item \textbf{Scaffolding produces large and reliable gains} (Aspect~0). The mean uplift over the GPT-5.4-mini no-scaffold baseline is $+\vmeanUpliftGpt{}$; \vfracBeatBase{}\ of runs and all \vnCells{}\ builder configurations exceed the baseline. The best scaffold reaches $\vbestRunAvg{}$ ($+\vbestRunUplift{}$), surpassing the Gemini-3.5-flash no-scaffold baseline on multiple tasks and approaching the human-inspired harness reference on the same backbone ($\vuserHarnessGptMacro{}$), despite using no human ToM-specific engineering.

  \item \textbf{The procedure is reproducible, though not deterministic} (Aspect~1). The mean within-cell standard deviation is $\vmeanWithinSd{}$, roughly an order of magnitude smaller than the main uplift. The remaining variance is concentrated in deterministic-solver strategies, where a single implementation error can substantially affect one benchmark.

  \item \textbf{The method is validation-efficient} (Aspect~2). Builders use a median of \vmedianNVal{}\ validation evaluations, show little evidence of overfitting to the $5\%$ validation slice (mean validation--full gap $\vmeanValFullGap{}$), and obtain no clear benefit from additional probing ($r=\vcorrNvalFull{}$). Builder quality matters more than validation budget.

  \item \textbf{The core mechanism is cognitive-load reduction} (Aspects~3, 7, and~8). Accuracy is strongly associated with the fraction of items answered by deterministic codes, rules and scaffolds ($r=0.72$). The best scaffolds combine a near-universal reliability floor, including format enforcement, routing, and greedy decoding, with higher-value task-structure exploitation, such as polarity logic, structured extraction, and deterministic solving.

  \item \textbf{Builder capability and effort dominate platform choice} (Aspects~4 and~6). Platform effects are second-order and conditional, whereas builder identity and reasoning effort are primary drivers of scaffold quality. For Opus-4.7, performance often improves monotonically with reasoning effort (Spearman $\rho=\veffortSpearman{}$).

  \item \textbf{The target model matters through headroom} (Aspect~5). Scaffolding helps most when the target leaves correctable errors on the table. As the target becomes stronger, the available headroom shrinks, and scaffolding must be applied more selectively to avoid disturbing or breaking those already-correct behaviors.

  \item \textbf{Remaining errors mark the less-compilable core of benchmark} (Aspect~9). Residual failures concentrate in deep belief recursion under deception and Bayesian goal inference, where reasoning cannot yet be fully reduced to rules or skills. Even so, top scaffolds can still fix about \vmeanFracFixed{}\ of baseline errors.
\end{itemize}
\vspace{4mm}
\textbf{Takeaway.} Strong-to-weak scaffolding works because a capable builder can act as a \emph{compiler of task competence}. It spends a one-time reasoning budget to identify structure in the task and encode that structure into an inference-time scaffold. Once compiled, a weaker and cheaper target model can execute the task at a level closer to that of much stronger models. This substitution is strongest for structured sub-problems whose decision procedures can be made explicit, but weaker for the genuinely hard core of benchmark reasoning. In ToM tasks, this includes nested belief tracking, deception, and counterfactual goal inference. Thus, scaffolding does not replace raw reasoning capability; it reallocates it. The builder performs the structural reasoning once, and the target model handles the residual cases that remain model-dependent.

Practically, the results suggest a simple recipe: use the strongest available builder, allocate high reasoning effort during scaffold construction, spend only a modest number of validation evaluations, prioritize cognitive offloading for provable sub-tasks, and, when budget permits, build several independent scaffolds and select or ensemble them to capture complementary repairs.

\section{Discussion and Future Work}
\label{sec:discussion}

\textbf{Benchmark selection.}
We use ToM benchmarks as a representative testbed rather than as the only setting where strong-to-weak scaffolding should apply. These benchmarks are well studied, contain diverse question types, and span a wide range of difficulty. This mixture is important for our purposes: some items require genuinely hard reasoning, while others expose regularities that a scaffold can exploit to reduce the target model's cognitive load. We do not view such exploitable structure as ``cheating.'' Under a fixed instruction and validation-only setting, discovering these structures is itself part of the builder model's capability. A strong builder should be able to inspect the validation slice, infer which parts of the task are compilable, and convert those observations into reusable inference-time skills. In this sense, the ToM benchmarks are useful precisely because they contain both scaffoldable structure and residual reasoning difficulty. This combination makes builder behavior more diverse and allows us to distinguish shallow prompt optimization from genuine task-structure discovery. Future work should test  strong-to-weak scaffolding paradigm on broader benchmark families, especially those with different mixtures of symbolic structure, ambiguity, and open-ended reasoning.

\textbf{Harness self-evolution.}
Strong-to-weak scaffolding also provides an empirical lens for studying harness self-evolution. Modern agentic coding environments such as Claude Code, Codex, and Cursor already shape model behavior through tools, prompts, workflows, and evaluation loops, but their design is still largely human-driven. Our setting asks whether models can automatically improve the harness around a target model, rather than merely produce answers within a fixed harness. This distinction is important: as agent systems become more capable, progress may come not only from improving the agent itself, but also from improving the environment that structures the agent's reasoning. By analyzing the scaffolds that builders create, we can identify which harness features consistently help, which add little value, and how different builders adapt the harness to the optimization target. In this way, automatic scaffold construction can inform human harness design while also pointing toward a future in which agents and their infrastructure co-evolve.

\textbf{Toward scaffolding as a benchmark.}
The strong-to-weak scaffolding setup can also be developed into a standard benchmark for builder models. A benchmark could provide a workspace, a weak target model, a fixed downstream task, and a validation set. The builder would be asked to write and refine a scaffold using only the validation data, and the final score would be measured on a hidden full test set using the target model. This would evaluate a capability that ordinary task benchmarks often miss: not whether the builder can answer the questions directly, but whether it can improve another model's ability to answer them. Secondary metrics could include validation usage, inference cost, scaffold complexity, code length, robustness across repeats, and the fraction of work offloaded from the target model. Such a benchmark would make harness design measurable in a free-form yet rigorous way, capturing both final accuracy and efficiency of the builder's solution.

\textbf{Two complementary routes to stronger systems.}
At a high level, there are two ways to improve performance on a task. One is to improve the model's internal capability; the other is to make the task easier for the model to execute. Most post-training work belongs to the first route, while harness and scaffold design belong to the second. Our work focuses on the second route, showing that a strong builder can partially substitute for target-model capability by reshaping the inference problem. These two routes should not be viewed as competitors. In the long run, models may be trained to use particular harnesses more effectively, while harnesses may be automatically optimized around the strengths and weaknesses of particular models. Recent work on using agents to design training recipes is one example of this interaction: the harness can guide model improvement, and improved models can in turn design better harnesses. Strong-to-weak scaffolding is therefore a step toward studying this broader co-evolution of models and inference environments. It suggests that future progress will depend not only on building stronger models, but also on learning how to structure tasks so that available models can deploy their capabilities more reliably.

%% file: sections/appendix.tex
\clearpage
\appendix
\section*{Appendix}

\section{Instruction Details}

In our main scaffold-building setting, we first place an instruction file in the workspace. The builder model then uses this file to construct the scaffold according to the provided instructions. The contents of the instruction file are described in detail below.

\begin{tcolorbox}[
  enhanced,
  breakable,
  width=0.98\linewidth,
  colback=tmlrPanel,
  colframe=tmlrBlue,
  boxrule=1.2pt,
  arc=6pt,
  left=5pt,right=5pt,top=4pt,bottom=2pt,
  title={\small Initial Instruction Content},
  coltitle=white,
  colbacktitle=tmlrBlueDark,
  fonttitle=\bfseries,
]
\small
\begin{lstlisting}[style=jsonTiny]
# Task: Build a Scaffold to Improve Target Model Benchmark Performance

You are entering a coding competition. Your goal is to build a scaffold/harness that improves the performance of a downstream tested model on a hidden benchmark set. You have access to a small validation set, which is a random 2% sample of the full benchmark. You should inspect the validation cases, design a scaffold, test it, analyze failures, and recursively refine your scaffold before the final full-set evaluation.

## Goal and Scoring

Your final grade is based on:

1. **Average performance across the four benchmarks** -- primary criterion.
2. **Number of validation evaluations used** -- secondary criterion.

Each validation evaluation costs 1. Use validation runs carefully. Do not repeatedly test small or unprincipled changes. Your scaffold should improve the target model's task performance and reduce its cognitive load.

## Workspace

You are assigned a workspace:

```text
/path/to/workspace
````

All code, outputs, logs, results, and notes must be saved under this workspace. You may not access other folders or search the internet.

The workspace already contains:

```text
validation.jsonl
engines.py
```

The full benchmark file will have the same format as `validation.jsonl`.

## Target Model

The tested model is:

```text
GPT-5.4-mini/Gemini-3.5-Flash
```

Please refer to `engines.py` for how to call the model.

## Required Files

You must maintain these two files throughout the process:

### 1. `performance.csv`

Record every validation evaluation. It must include four columns for the four benchmark scores.

Suggested format:

```csv
run_id,split,benchmark_1,benchmark_2,benchmark_3,benchmark_4,average,notes
```

### 2. `optimization.md`

Keep a clear recursive refinement log. For each iteration, record:

* What you inspected
* What scaffold change you made
* Why you made it
* Validation result
* What you learned
* Next planned improvement

## Development Process

1. Inspect the workspace, especially `validation.jsonl` and `engines.py`.
2. Understand the input/output format, benchmark types, and evaluation requirements.
3. Build an initial scaffold and baseline evaluation script.
4. Run validation only after meaningful scaffold changes.
5. Analyze failures and refine the scaffold recursively.
6. Stop when performance plateaus or further changes risk overfitting.
7. Prepare the final full-set evaluation script.

Your scaffold may include prompt templates, task routing, answer-format enforcement, few-shot examples, verification steps, deterministic preprocessing/postprocessing, or other methods that help the target model perform better.

Avoid hard-coding validation answers. No cheating. Optimize for generalization to the hidden full set.

## Final Deliverable

Create a bash script:

```text
full_run.sh
```

The evaluator should only need to insert or pass the full-set path to run your final harness.

Recommended usage:

```bash
bash full_run.sh /path/to/full_set.jsonl
```

The script should:

1. Load the given full-set file
2. Run your final scaffold on it
3. Save predictions/results inside the workspace
4. Append the full-set benchmark results to `performance.csv`
5. Full set testing should be parallel call of target models with max worker equals 16
6. Full set final results or analysis output should be saved in independent folder `final_eval`

## Final Reminder

You have only one final chance on the hidden full set. Be strategic: inspect carefully, make each validation run count, recursively improve the scaffold, and prioritize robust benchmark-wide performance over validation overfitting.
\end{lstlisting}
\end{tcolorbox}

\section{Complete Per-Run Results}
\label{app:allruns}

The table below lists all \vnRunsTotal{}\unskip\ runs, including their factor coordinates, graded per-benchmark accuracies, macro-average accuracies, and the number of validation evaluations used by the builder.

\begin{table*}[h]
\centering
\caption{All runs, sorted by target / builder / platform / repeat. ``Avg.'' is the macro average of the four benchmark accuracies (primary metric). ``Val.'' is the number of validation evaluations used.}
\label{tab:appendix}
\setlength{\tabcolsep}{3pt}
\renewcommand{\arraystretch}{1.05}
\small
\setlength{\tabcolsep}{2pt}
\resizebox{\textwidth}{!}{\tiny\input{tables/appendix_allruns_1}}
\end{table*}

\begin{table*}[h]
\centering
% \caption{All runs, sorted by target / builder / platform / repeat. ``Avg.'' is the macro average of the four benchmark accuracies (primary metric). ``Val.'' is the number of validation evaluations used.}
\label{tab:appendix}
\setlength{\tabcolsep}{3pt}
\renewcommand{\arraystretch}{1.05}
\small
\setlength{\tabcolsep}{2pt}
\resizebox{\textwidth}{!}{\tiny\input{tables/appendix_allruns_2}}
\end{table*}

%% file: tables/appendix_allruns_1.tex
\resizebox{\linewidth}{!}{
\begin{tabular}{llllccccccc}
\toprule
Run & Platform & Builder & Target & Rep & BigToM & Hi-ToM & MMToM & MuMA & Avg. & Val. \\
\midrule
gemini35flash-gemini35flash & Cursor & Gemini-3.5-flash & gemini-3.5-flash & 1 & 0.874 & 0.738 & 0.770 & 0.990 & \textbf{0.843} & 4 \\
gemini35flash-gemini35flash-2 & Cursor & Gemini-3.5-flash & gemini-3.5-flash & 2 & 1.000 & 0.699 & 0.720 & 0.916 & \textbf{0.834} & 3 \\
gemini35flash-gemini35flash-3 & Cursor & Gemini-3.5-flash & gemini-3.5-flash & 3 & 0.971 & 0.816 & 0.977 & 0.991 & \textbf{0.939} & 2 \\
gemini31pro-gemini35flash & Cursor & Gemini-3.1-Pro & gemini-3.5-flash & 1 & 0.870 & 0.697 & 0.772 & 0.959 & \textbf{0.825} & 4 \\
gemini31pro-gemini35flash-2 & Cursor & Gemini-3.1-Pro & gemini-3.5-flash & 2 & 0.899 & 0.818 & 0.933 & 0.947 & \textbf{0.899} & 5 \\
gemini31pro-15544413-3 & Cursor & Gemini-3.1-Pro & gemini-3.5-flash & 3 & 0.914 & 0.832 & 0.948 & 0.978 & \textbf{0.918} & 25 \\
gpt55-gemini35flash & Cursor & GPT-5.5 & gemini-3.5-flash & 1 & 1.000 & 0.848 & 0.768 & 0.934 & \textbf{0.888} & 6 \\
gpt55-gemini35flash-2 & Cursor & GPT-5.5 & gemini-3.5-flash & 2 & 1.000 & 0.824 & 0.940 & 0.908 & \textbf{0.918} & 3 \\
gpt55-gemini35flash-3 & Cursor & GPT-5.5 & gemini-3.5-flash & 3 & 1.000 & 0.824 & 0.883 & 0.880 & \textbf{0.897} & 3 \\
\bottomrule
\end{tabular}
}

%% file: tables/appendix_allruns_2.tex
\resizebox{\linewidth}{!}{
\begin{tabular}{llllccccccc}
\toprule
% Run & Platform & Builder & Target & Rep & BigToM & Hi-ToM & MMToM & MuMA & Avg. & Val. \\
% \midrule
% gemini35flash-gemini35flash & Cursor & Gemini-3.5-flash & gemini-3.5-flash & 1 & 0.874 & 0.738 & 0.770 & 0.990 & \textbf{0.843} & 4 \\
% gemini35flash-gemini35flash-2 & Cursor & Gemini-3.5-flash & gemini-3.5-flash & 2 & 1.000 & 0.699 & 0.720 & 0.916 & \textbf{0.834} & 3 \\
% gemini35flash-gemini35flash-3 & Cursor & Gemini-3.5-flash & gemini-3.5-flash & 3 & 0.971 & 0.816 & 0.977 & 0.991 & \textbf{0.939} & 2 \\
% gemini31pro-gemini35flash & Cursor & Gemini-3.1-Pro & gemini-3.5-flash & 1 & 0.870 & 0.697 & 0.772 & 0.959 & \textbf{0.825} & 4 \\
% gemini31pro-gemini35flash-2 & Cursor & Gemini-3.1-Pro & gemini-3.5-flash & 2 & 0.899 & 0.818 & 0.933 & 0.947 & \textbf{0.899} & 5 \\
% gemini31pro-15544413-3 & Cursor & Gemini-3.1-Pro & gemini-3.5-flash & 3 & 0.914 & 0.832 & 0.948 & 0.978 & \textbf{0.918} & 25 \\
% gpt55-gemini35flash & Cursor & GPT-5.5 & gemini-3.5-flash & 1 & 1.000 & 0.848 & 0.768 & 0.934 & \textbf{0.888} & 6 \\
% gpt55-gemini35flash-2 & Cursor & GPT-5.5 & gemini-3.5-flash & 2 & 1.000 & 0.824 & 0.940 & 0.908 & \textbf{0.918} & 3 \\
% gpt55-gemini35flash-3 & Cursor & GPT-5.5 & gemini-3.5-flash & 3 & 1.000 & 0.824 & 0.883 & 0.880 & \textbf{0.897} & 3 \\
grok01-gemini35flash & Cursor & Grok-0.1 & gemini-3.5-flash & 1 & 0.876 & 0.813 & 0.813 & 0.971 & \textbf{0.868} & 2 \\
grok01-gemini35flash-2 & Cursor & Grok-0.1 & gemini-3.5-flash & 2 & 0.513 & 0.634 & 0.640 & 0.952 & \textbf{0.685} & 3 \\
grok01-gemini35flash-3 & Cursor & Grok-0.1 & gemini-3.5-flash & 3 & 0.896 & 0.651 & 0.637 & 0.969 & \textbf{0.788} & 3 \\
opus47-gemini35flash & Cursor & Opus-4.7 (x-high) & gemini-3.5-flash & 1 & 0.981 & 0.817 & 0.943 & 0.986 & \textbf{0.932} & 2 \\
opus47-gemini35flash-2 & Cursor & Opus-4.7 (x-high) & gemini-3.5-flash & 2 & 0.985 & 0.775 & 0.978 & 0.992 & \textbf{0.933} & 3 \\
opus47-gemini35flash-3 & Cursor & Opus-4.7 (x-high) & gemini-3.5-flash & 3 & 0.959 & 0.814 & 0.873 & 0.969 & \textbf{0.904} & 4 \\
codex53-gpt54mini & Cursor & Codex-5.3 & gpt-5.4-mini & 1 & 1.000 & 0.587 & 0.367 & 0.518 & \textbf{0.618} & 4 \\
codex53-gpt54mini-2 & Cursor & Codex-5.3 & gpt-5.4-mini & 2 & 1.000 & 0.552 & 0.598 & 0.506 & \textbf{0.664} & 4 \\
codex53-gpt54mini-3 & Cursor & Codex-5.3 & gpt-5.4-mini & 3 & 0.900 & 0.804 & 0.358 & 0.498 & \textbf{0.640} & 7 \\
gptcodex-codex53-gpt54mini & GPT Codex & Codex-5.3 & gpt-5.4-mini & 1 & 1.000 & 0.560 & 0.762 & 0.541 & \textbf{0.716} & 4 \\
gptcodex-codex53-gpt54mini-2 & GPT Codex & Codex-5.3 & gpt-5.4-mini & 2 & 1.000 & 0.568 & 0.597 & 0.501 & \textbf{0.667} & 2 \\
gptcodex-codex53-gpt54mini-3 & GPT Codex & Codex-5.3 & gpt-5.4-mini & 3 & 1.000 & 0.682 & 0.695 & 0.606 & \textbf{0.746} & 5 \\
gemini35flash-gpt54mini & Cursor & Gemini-3.5-flash & gpt-5.4-mini & 1 & 0.958 & 0.665 & 0.697 & 0.670 & \textbf{0.747} & 4 \\
gemini35flash-gpt54mini-2 & Cursor & Gemini-3.5-flash & gpt-5.4-mini & 2 & 1.000 & 0.731 & 0.790 & 0.844 & \textbf{0.841} & 3 \\
gemini35flash-gpt54mini-3 & Cursor & Gemini-3.5-flash & gpt-5.4-mini & 3 & 1.000 & 0.739 & 0.847 & 0.817 & \textbf{0.851} & 5 \\
gemini31pro-gpt54mini & Cursor & Gemini-3.1-Pro & gpt-5.4-mini & 1 & 0.926 & 0.682 & 0.518 & 0.581 & \textbf{0.677} & 4 \\
gemini31pro-gpt54mini-2 & Cursor & Gemini-3.1-Pro & gpt-5.4-mini & 2 & 0.932 & 0.776 & 0.657 & 0.599 & \textbf{0.741} & 5 \\
gemini31pro-gpt54mini-3 & Cursor & Gemini-3.1-Pro & gpt-5.4-mini & 3 & 0.872 & 0.738 & 0.680 & 0.600 & \textbf{0.722} & 11 \\
gpt54mini-gpt54mini & Cursor & GPT-5.4-mini & gpt-5.4-mini & 1 & 1.000 & 0.831 & 0.803 & 0.414 & \textbf{0.762} & 6 \\
gpt54mini-gpt54mini-2 & Cursor & GPT-5.4-mini & gpt-5.4-mini & 2 & 1.000 & 0.643 & 0.560 & 0.607 & \textbf{0.702} & 4 \\
gpt54mini-gpt54mini-3 & Cursor & GPT-5.4-mini & gpt-5.4-mini & 3 & 0.883 & 0.575 & 0.682 & 0.468 & \textbf{0.652} & 7 \\
gptcodex-gpt54mini-gpt54mini & GPT Codex & GPT-5.4-mini & gpt-5.4-mini & 1 & 1.000 & 0.556 & 0.447 & 0.454 & \textbf{0.614} & 6 \\
gptcodex-gpt54mini-gpt54mini-2 & GPT Codex & GPT-5.4-mini & gpt-5.4-mini & 2 & 1.000 & 0.571 & 0.400 & 0.450 & \textbf{0.605} & 3 \\
gptcodex-gpt54mini-gpt54mini-3 & GPT Codex & GPT-5.4-mini & gpt-5.4-mini & 3 & 1.000 & 0.720 & 0.823 & 0.452 & \textbf{0.749} & 2 \\
gpt55-gpt54mini & Cursor & GPT-5.5 & gpt-5.4-mini & 1 & 1.000 & 0.831 & 0.597 & 0.921 & \textbf{0.837} & 15 \\
gpt55-gpt54mini-2 & Cursor & GPT-5.5 & gpt-5.4-mini & 2 & 1.000 & 0.831 & 0.873 & 0.928 & \textbf{0.908} & 3 \\
gpt55-gpt54mini-3 & Cursor & GPT-5.5 & gpt-5.4-mini & 3 & 1.000 & 0.824 & 0.845 & 0.612 & \textbf{0.820} & 3 \\
gptcodex-gpt55-gpt54mini & GPT Codex & GPT-5.5 & gpt-5.4-mini & 1 & 1.000 & 0.843 & 0.888 & 0.916 & \textbf{0.912} & 7 \\
gptcodex-gpt55-gpt54mini-2 & GPT Codex & GPT-5.5 & gpt-5.4-mini & 2 & 1.000 & 0.660 & 0.912 & 0.918 & \textbf{0.872} & 2 \\
gptcodex-gpt55-gpt54mini-3 & GPT Codex & GPT-5.5 & gpt-5.4-mini & 3 & 1.000 & 0.828 & 0.937 & 0.848 & \textbf{0.903} & 5 \\
grok01-gpt54mini & Cursor & Grok-0.1 & gpt-5.4-mini & 1 & 0.496 & 0.623 & 0.593 & 0.596 & \textbf{0.577} & 2 \\
grok01-gpt54mini-2 & Cursor & Grok-0.1 & gpt-5.4-mini & 2 & 0.492 & 0.620 & 0.487 & 0.459 & \textbf{0.514} & 3 \\
grok01-gpt54mini-3 & Cursor & Grok-0.1 & gpt-5.4-mini & 3 & 0.852 & 0.534 & 0.530 & 0.479 & \textbf{0.599} & 3 \\
claudecode-opus47-gpt54mini & Claude Code & Opus-4.7 (x-high) & gpt-5.4-mini & 1 & 0.991 & 0.785 & 0.830 & 0.880 & \textbf{0.871} & 6 \\
claudecode-opus47-gpt54mini-2 & Claude Code & Opus-4.7 (x-high) & gpt-5.4-mini & 2 & 0.996 & 0.797 & 0.775 & 0.900 & \textbf{0.867} & 6 \\
claudecode-opus47-gpt54mini-3 & Claude Code & Opus-4.7 (x-high) & gpt-5.4-mini & 3 & 1.000 & 0.782 & 0.787 & 0.947 & \textbf{0.879} & 7 \\
opus47-gpt54mini & Cursor & Opus-4.7 (x-high) & gpt-5.4-mini & 1 & 0.902 & 0.722 & 0.905 & 0.822 & \textbf{0.838} & 5 \\
opus47-gpt54mini-2 & Cursor & Opus-4.7 (x-high) & gpt-5.4-mini & 2 & 1.000 & 0.916 & 0.728 & 0.818 & \textbf{0.865} & 5 \\
opus47-gpt54mini-3 & Cursor & Opus-4.7 (x-high) & gpt-5.4-mini & 3 & 0.929 & 0.747 & 0.703 & 0.890 & \textbf{0.817} & 4 \\
claudecode-opus47high-gpt54mini & Claude Code & Opus-4.7 (high) & gpt-5.4-mini & 1 & 0.966 & 0.782 & 0.688 & 0.940 & \textbf{0.844} & 4 \\
claudecode-opus47high-gpt54mini-2 & Claude Code & Opus-4.7 (high) & gpt-5.4-mini & 2 & 0.967 & 0.726 & 0.738 & 0.784 & \textbf{0.804} & 6 \\
claudecode-opus47high-gpt54mini-3 & Claude Code & Opus-4.7 (high) & gpt-5.4-mini & 3 & 1.000 & 0.743 & 0.818 & 0.762 & \textbf{0.831} & 3 \\
opus47high-gpt54mini & Cursor & Opus-4.7 (high) & gpt-5.4-mini & 1 & 0.877 & 0.659 & 0.772 & 0.656 & \textbf{0.741} & 5 \\
opus47high-gpt54mini-2 & Cursor & Opus-4.7 (high) & gpt-5.4-mini & 2 & 0.835 & 0.771 & 0.812 & 0.851 & \textbf{0.817} & 4 \\
opus47high-gpt54mini-3 & Cursor & Opus-4.7 (high) & gpt-5.4-mini & 3 & 0.891 & 0.751 & 0.832 & 0.750 & \textbf{0.806} & 5 \\
claudecode-opus47low-gpt54mini & Claude Code & Opus-4.7 (low) & gpt-5.4-mini & 1 & 0.869 & 0.709 & 0.647 & 0.571 & \textbf{0.699} & 2 \\
claudecode-opus47low-gpt54mini-2 & Claude Code & Opus-4.7 (low) & gpt-5.4-mini & 2 & 0.880 & 0.729 & 0.513 & 0.634 & \textbf{0.689} & 4 \\
claudecode-opus47low-gpt54mini-3 & Claude Code & Opus-4.7 (low) & gpt-5.4-mini & 3 & 0.873 & 0.623 & 0.685 & 0.591 & \textbf{0.693} & 2 \\
opus47low-gpt54mini & Cursor & Opus-4.7 (low) & gpt-5.4-mini & 1 & 0.872 & 0.738 & 0.667 & 0.836 & \textbf{0.778} & 4 \\
opus47low-gpt54mini-2 & Cursor & Opus-4.7 (low) & gpt-5.4-mini & 2 & 0.866 & 0.677 & 0.630 & 0.613 & \textbf{0.697} & 3 \\
opus47low-gpt54mini-3 & Cursor & Opus-4.7 (low) & gpt-5.4-mini & 3 & 0.963 & 0.652 & 0.512 & 0.708 & \textbf{0.708} & 4 \\
claudecode-opus47medium-gpt54mini & Claude Code & Opus-4.7 (med) & gpt-5.4-mini & 1 & 0.859 & 0.703 & 0.635 & 0.888 & \textbf{0.771} & 5 \\
claudecode-opus47medium-gpt54mini-2 & Claude Code & Opus-4.7 (med) & gpt-5.4-mini & 2 & 0.968 & 0.705 & 0.783 & 0.792 & \textbf{0.812} & 6 \\
claudecode-opus47medium-gpt54mini-3 & Claude Code & Opus-4.7 (med) & gpt-5.4-mini & 3 & 0.945 & 0.748 & 0.823 & 0.937 & \textbf{0.863} & 5 \\
opus47medium-gpt54mini & Cursor & Opus-4.7 (med) & gpt-5.4-mini & 1 & 1.000 & 0.762 & 0.693 & 0.738 & \textbf{0.798} & 2 \\
opus47medium-gpt54mini-2 & Cursor & Opus-4.7 (med) & gpt-5.4-mini & 2 & 0.893 & 0.566 & 0.733 & 0.464 & \textbf{0.664} & 7 \\
opus47medium-gpt54mini-3 & Cursor & Opus-4.7 (med) & gpt-5.4-mini & 3 & 1.000 & 0.710 & 0.837 & 0.847 & \textbf{0.848} & 6 \\
claudecode-sonnet46-gpt54mini & Claude Code & Sonnet-4.6 & gpt-5.4-mini & 1 & 0.994 & 0.586 & 0.598 & 0.499 & \textbf{0.669} & 11 \\
claudecode-sonnet46-gpt54mini-2 & Claude Code & Sonnet-4.6 & gpt-5.4-mini & 2 & 1.000 & 0.700 & 0.830 & 0.950 & \textbf{0.870} & 8 \\
claudecode-sonnet46-gpt54mini-3 & Claude Code & Sonnet-4.6 & gpt-5.4-mini & 3 & 0.962 & 0.731 & 0.793 & 0.830 & \textbf{0.829} & 4 \\
sonnet46-gpt54mini & Cursor & Sonnet-4.6 & gpt-5.4-mini & 1 & 0.981 & 0.729 & 0.642 & 0.810 & \textbf{0.790} & 8 \\
sonnet46-gpt54mini-2 & Cursor & Sonnet-4.6 & gpt-5.4-mini & 2 & 0.968 & 0.763 & 0.850 & 0.912 & \textbf{0.873} & 5 \\
sonnet46-gpt54mini-3 & Cursor & Sonnet-4.6 & gpt-5.4-mini & 3 & 0.959 & 0.762 & 0.737 & 0.858 & \textbf{0.829} & 5 \\
\bottomrule
\end{tabular}
}